\documentclass[conference]{IEEEtran}

\usepackage{float}
\usepackage{xcolor}
\usepackage{pgffor}
\usepackage{authblk}
\usepackage{graphicx}
\usepackage{amsmath}
\usepackage{mathtools}
\usepackage{hyperref}
\usepackage{etoolbox}
\usepackage{tabularx}
\usepackage{xltabular}
\usepackage{booktabs}
\usepackage{subcaption}
\usepackage[T1]{fontenc}
\usepackage[utf8]{inputenc}
\usepackage[super]{nth}
\usepackage[english]{babel}
\usepackage[autostyle=true]{csquotes}
\usepackage[nolist, nohyperlinks]{acronym}
\usepackage[capitalize, sort, english]{cleveref}

\newcolumntype{Y}{>{\raggedleft\arraybackslash}X}
\newcolumntype{R}{>{\hsize=0.45\hsize}Y}

\newcommand{\artdl}{ArtDL}
\newcommand{\iconart}{IconArt}

\newsavebox{\imgbox}
\newcommand{\imagemap}[2][4.62cm]{%
  \sbox{\imgbox}{\includegraphics[height=#1]{#2}}%
  \begin{subfigure}[t]{\wd\imgbox}
    \centering
    \usebox{\imgbox}
    \caption{}
  \end{subfigure}%
}

\begin{acronym}
\acro{AI}{\textit{Artificial Intelligence}}
\acro{CAM}{\textit{Class Activation Map}}
\acro{CLIP}{\textit{Contrastive Language--Image Pre-training}}
\acro{DNN}{\textit{Deep Neural Network}}
\acro{GLAM}{\textit{Galleries, Libraries, Archives, and Museums}}
\acro{IoU}{\textit{Intersection over Union}}
\acro{MICE}{\textit{Multivariate Imputation by Chained Equations}}
\acro{VLM}{\textit{Vision--Language Model}}
\acro{ViT}{\textit{Vision Transformer}}
\acro{XAI}{\textit{Explainable Artificial Intelligence}}
\end{acronym}

\usepackage[%
    backend=biber,
    style=authoryear,
    natbib=true,
    sorting=nyt,
    uniquelist=false,
    uniquename=false,
    maxbibnames=99,
    maxcitenames=2,
]{biblatex}

\title{On the Explainability of Vision--Language \\Models in Art History}

\author[1]{Stefanie Schneider}

\affil[1]{Marburg University, Marburg, Germany\thanks{Corresponding author: \href{mailto:stefanie.schneider@uni-marburg.de}{stefanie.schneider@uni-marburg.de}}}

\begin{document}

\maketitle

\begin{abstract}
\acp{VLM} transfer visual and textual data into a shared embedding space.
In so doing, they enable a wide range of multimodal tasks, while also raising critical questions about the nature of machine \enquote*{understanding.}
In this paper, we examine how \ac{XAI} methods can render the visual reasoning of a \ac{VLM}---namely, \acs{CLIP}---legible in art-historical contexts.
To this end, we evaluate seven methods, combining zero-shot localization experiments with human interpretability studies.
Our results indicate that, while these methods capture some aspects of human interpretation, their effectiveness hinges on the conceptual stability and representational availability of the examined categories.
\end{abstract}

\acresetall


\section{Introduction}
\label{sec:introduction}

\acp{VLM} have, in recent years, become remarkably versatile instruments of analysis.
By aligning visual and linguistic information within a shared embedding space, they can perform a wide range of multimodal tasks---from retrieval \citep{RadfordKHRGASAM21} and captioning \citep{liLSH23} to zero-shot classification \citep{zhaiWMSK0B22}.
However, this versatility has made them the object of sustained criticism: not only due to the opacity of their internal mechanisms, but also because of the ethical, sociotechnical, and epistemological assumptions encoded in their design.
It has been questioned what forms of \enquote*{understanding} such models enact, how their embeddings reify social hierarchies, and to which extent their apparent generality conceals dependencies on biased, uncurated data \citep{BhallaOSCL24, birhanePK2021, benderGMS21}.
In short, we might ask: what does it mean for a model to \textit{see}?

\begin{figure}[t!]
    \centering
    \begin{subfigure}[t]{0.487\columnwidth}
        \includegraphics[width=\textwidth]{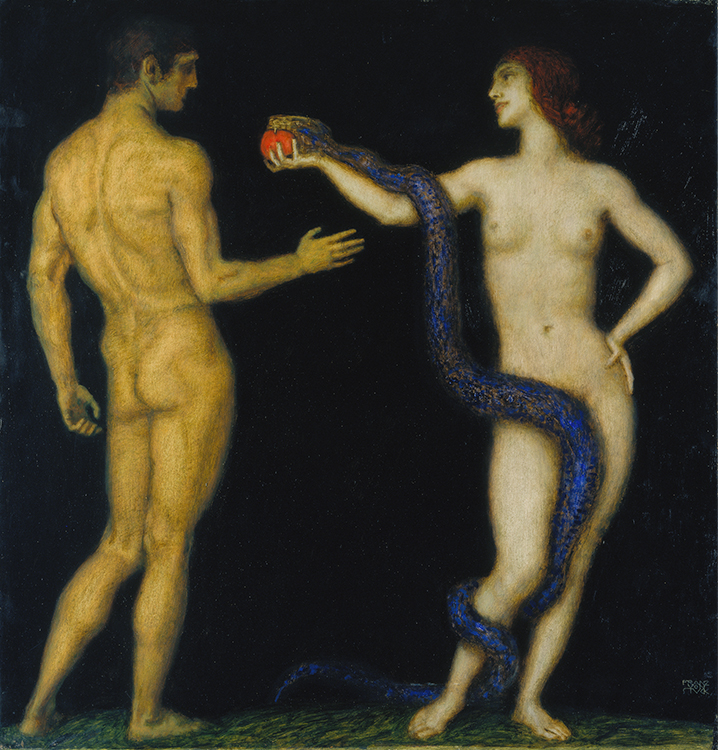}
        \caption{Original Image}
    \end{subfigure}%
    \hfill
    \begin{subfigure}[t]{0.487\columnwidth}
        \includegraphics[width=\textwidth]{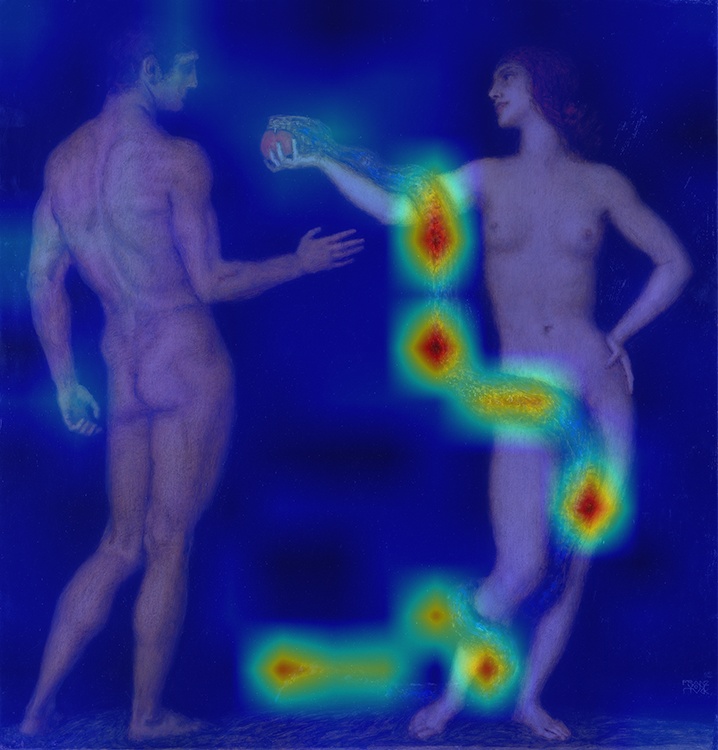}
        \caption{Saliency Map}
    \end{subfigure}

    \caption{
        The saliency map highlights, in \textit{red}, the image regions most strongly associated with the concept of the \enquote{snake} in Franz von Stuck's \textit{Adam and Eve} (c.~1920).
    }
    \label{fig:saliency-example}
\end{figure}

This question is particularly relevant in fields where visual meaning is historically and semantically dense---where \enquote*{objects,} in the broadest sense, cannot be reduced to mere labels or descriptive tokens.
Art history is exemplary in this regard: here, the visual is not simply perceived, but interpreted through culturally sedimented conventions of style, iconography, and material practice.
Nevertheless, models such as \acsu{CLIP} (\acl{CLIP}; \cite{RadfordKHRGASAM21}) are now routinely employed \enquote*{out of the box} for art-historical retrieval and analysis on digital platforms \citep{springsteinSRHK21, offertB2023}, often without a clear understanding of which kinds of visual concepts---formal, iconographic, or affective---are encoded in their embeddings.
As \citet{birhanePK2021} demonstrate, the datasets on which these models are based (e.g., LAION-400M, \cite{schuhmannVBKM2021}) contain structural biases, non-consensual imagery, and stereotypical or pornographic representations that reflect the discriminatory nature of the web rather than a neutral visual commons.
The epistemic opacity of \acp{VLM} thus becomes a methodological issue: how can search results be interpreted when the model itself embeds an unacknowledged theory of vision?
\ac{CLIP}, in particular, epitomizes the multimodal turn as a large-scale \ac{VLM} trained on millions of image–text pairs scraped from the web.
Its embedding space constitutes not only a technical geometry of similarity but what \citet{impettO2022} call a \enquote{vector imaginary} of contemporary visual culture---a statistical condensation of what is collectively pictured and named online.
This makes \ac{CLIP} both uniquely powerful and uniquely problematic for art-historical inquiry.
On the one hand, its zero-shot capacity enables the retrieval of artworks that are stylistically or iconographically related without supervision.
Yet the same mechanism also perpetuates the omissions of its training corpus, reproducing a visuality that is historically and culturally uneven.
Against this backdrop, we ask a central question: to what extent can \ac{XAI} methods render the visual logic of \ac{CLIP} legible to human interpreters, thereby strengthening the methodological robustness of \acp{VLM} in art-historical contexts?
To explore this, we comparatively evaluate seven \ac{XAI} methods spanning three paradigms: 
(1)~gradient-based methods that backpropagate class-specific gradients into feature maps (Grad-CAM, \cite{gradcam}, 2017; Grad-CAM++, \cite{gradcam++}; LayerCAM, \cite{layercam}; LeGrad, \cite{legrad});
(2)~score-based, gradient-free methods that measure the influence of image regions on the model-predicted score (ScoreCAM, \cite{scorecam}; gScoreCAM, \cite{gscorecam});
and (3)~\ac{CLIP}-specific approaches that intervene directly in the inference pipeline (CLIP Surgery, \cite{clip-surgery}).
Each of these methods generates, for a given text prompt, a saliency map that visualizes the contribution of an image region to the model's predicted score (\Cref{fig:saliency-example}).
In this paper, our objective is not just to identify the most effective technique, but also to explore the boundaries of explainability under conditions of zero-shot inference and domain transfer.
We, thus, include methods that, while no longer state-of-the-art in some cases, remain prevalent in practice, in order to highlight the interpretive and subjective dimensions of \enquote*{explainability} itself.

To examine these dimensions systematically, we adopt a two-stage evaluation framework.
First, a quantitative case study employs two art-historical datasets---\iconart\ \citep{gonthierGLB18} and \artdl\ \citep{MilaniF21}---to measure the localization accuracy of the methods under zero-shot conditions.
Then, in an online survey with participants trained in art history, the interpretability of these same methods is evaluated, situating the previously obtained results within the variability of human visual judgment.
Building on \citet{BhallaOSCL24}, who demonstrate that \ac{CLIP} embeddings can be decomposed into sparse, interpretable concept spaces, our studies ask whether such interpretability extends to visual explanations: that is, do \ac{XAI} methods disclose a model's internal conceptual structure---or merely aestheticize its opacity?
This question is especially pertinent given critiques by \citet{birhanePK2021, benderGMS21}, who argue that large-scale, web-scraped datasets can perpetuate hegemonic biases under the guise of generality.
If explainability techniques cannot illuminate these latent structures, they may reiterate rather than expose the ideological patterns of machine vision.
From this standpoint, we articulate three research questions:
(1)~\textit{How effectively do \ac{XAI} methods localize iconographic objects in artworks under zero-shot conditions without fine-tuning?}
This establishes a baseline: can methods trained on everyday imagery nonetheless delineate the complex, symbolically charged forms within artworks beyond their training distribution?
(2)~\textit{Does the visual relevance of these maps correspond to human judgments?}
If saliency maps claim to visualize \enquote*{what the model sees,} their validity must be tested against human perception---specifically against the art-historically informed gaze.
(3)~\textit{Which factors, such as object size and concept abstraction, drive performance differences?}
Here, we connect measurable attributes with semantic ones, aligning computational error analysis with questions of representation central to art-historical inquiry.
By testing how---and where---\ac{XAI} methods succeed or fail in making \ac{CLIP}'s mechanisms visible, we contribute to a broader methodological debate about how digital art history might critically engage with the epistemic structures of machine vision.
Our aim, in other words, is to determine not only what \acp{VLM} attend to in works of art, but examine how their patterns of attention either align---or fail to align---with human interpretive conventions.

The remainder of this paper is structured as follows:
In \cref{sec:method-selection}, we outline the rationale for selecting the examined \ac{XAI} techniques.
\Cref{sec:study-1} presents the first case study, which quantitatively evaluates localization accuracy using two art-historical datasets under zero-shot conditions.
\Cref{sec:study-2} turns to the second case study: an online survey that assesses the interpretability of saliency maps from an art-historical perspective.
\cref{sec:discussion} then synthesizes these findings, tracing their methodological implications for explainability, interpretability, and the critical analysis of machine vision within digital art history.

\section{Method Selection}
\label{sec:method-selection}

Since the advent of \ac{AI}, researchers have sought to make its internal mechanisms intelligible.
This imperative has only intensified further with the resurgence of \acp{DNN} in the 2000s---highly non-linear statistical models with millions of parameters.
The demand for explanation, however, predates deep learning; it can be traced back to the early attempts to understand expert systems \citep[e.g.,][]{buchananS1984}, which, emerging in the 1960s and 1970s, marked one of the first conceptual turns in \ac{AI} towards the explicability of machine reasoning \citep{coyB1987}.
Designed to mimic the problem-solving strategies of human specialists within narrowly defined use cases, these systems aimed not only to provide decisions, but also to explain the reasoning behind them \citep{harmonMM1989}.
Their architecture---typically comprising a knowledge base and an inference engine---explicitly separated domain-specific expertise from domain-independent reasoning procedures \citep{puppe1988}.
Although modern \acp{DNN} have largely abandoned such symbolic, rule-based representations in favor of statistical learning, the epistemological tension first articulated in the age of expert systems---between algorithmic performance and interpretability---remains a defining problem for contemporary research in \ac{XAI}.

We do not, however, in this section, intend to provide a historical or taxonomic overview of \ac{XAI} methods---an exercise already undertaken in recent years from a variety of perspectives \citep[e.g.,][]{speith22, saeedO23, hassijaCMSGHSSMH24}.
Rather, we motivate our decision to focus on a specific class of \textit{perceptive interpretability} methods, i.e., saliency-based visualization methods applied \textit{post hoc} to a pre-trained model, in our case \ac{CLIP}.
Our goal is not to revisit the broader epistemological debates surrounding explainability, but to examine how visual explanations---those that make a model's internal reasoning visible---operate in the specific and practically relevant context of \acp{VLM}.
In this context, perceptive interpretability methods, and saliency-based visualizations in particular, occupy a unique position at the interface of human perception and machine inference, as they generate intuitive, human-readable representations of model attention.
In doing so, they provide a spatial grammar through which one might then ask where a model \enquote*{looks} when associating textual prompts with visual content.
We therefore select methods according to three criteria:
(1)~the method must produce spatially localized, human-inspectable heatmaps over the input image;
(2)~it must operate \textit{post hoc} on \ac{CLIP}---without retraining or architectural modification---to ensure comparability across prompts and datasets; and
(3)~it must be either widely adopted as a baseline or specifically adapted to \ac{CLIP}'s dual-encoder architecture, making text–image interactions explicit.
We exclude approaches that introduce additional hyperparameters, such as CLIP-LIME \citep{kazmierczakBFF24}.
Likewise, attention-based methods \citep[e.g.,][]{abnarZ20} are omitted, as their reliance on attention weights has been shown to correlate only weakly with decision relevance \citep{jainW19}.
Also excluded are methods such as CLIP-Dissect \citep{clip-dissect}: although they yield valuable insights into the representational topology of \ac{CLIP}, they do not address the specific perceptual question that motivates our study, i.e., where in the image does the model locate the evidence for a given prompt?

Given these constraints, we have identified three methodological paradigms for evaluation:
\textit{gradient-based} methods trace how the model's internal signals---its gradients---shift in response to visual stimuli, revealing which regions of an image most strongly determine a given decision;
\textit{score-based}, gradient-free methods, by contrast, perturb the input: they mask regions of the image and measure the resulting change in the model's output, inferring from these changes which areas carry the greatest weight;
\textit{\ac{CLIP}-specific} interventions operate directly on the multimodal architecture, adjusting how \ac{CLIP} integrates textual and visual information to make visible the relational mechanics between words and image regions.
Within the gradient-based family, we include Grad-CAM \citep{gradcam} as the canonical baseline, Grad-CAM++ \citep{gradcam++} as its extension to multi-instance scenarios via higher-order gradient weighting, and LayerCAM \citep{layercam} as a refinement that pools activations from intermediate convolutional layers to enhance spatial fidelity.
LeGrad \citep{legrad} further optimizes this aggregation process, improving localization accuracy while reducing sensitivity to layer selection.
The second paradigm---score-based, gradient-free evaluation---is exemplified by ScoreCAM \citep{scorecam} and gScoreCAM \citep{gscorecam}.
Both rely on selective occlusion: by systematically masking portions of the image and recording how the model's scores change, they construct saliency maps that recompose these perturbations into a spatial account of the model's visual attention.
Finally, CLIP Surgery \citep{clip-surgery} represents a model-specific intervention at inference time.
By reparameterizing the forward pass and adapting Grad-CAM-like mechanisms to \ac{CLIP}'s dual-encoder architecture, it produces activation maps that more explicitly disentangle the textual and visual streams of the model---rendering visible, in effect, how the network's multimodal alignments are spatially instantiated.

However, while such techniques have become---at least to some extent---standard practice within machine vision, their epistemic foundations remain contested.
According to \citet{Miller19}, explanations are not static products but dialogical processes that unfold within particular social contexts: they presuppose both an \textit{explainer} and an \textit{explainee}, whose beliefs, expectations, and cognitive biases shape not only the form of an explanation, but also how adequate it is perceived to be.
Explanations, \citet{Miller19} emphasizes, are inherently contrastive (\textit{Why this rather than that?}), selective (\textit{Which of the many possible causes should be prioritized?}), and social (\textit{How should this explanation be shaped for the person to whom it is addressed?}).
In this sense, Miller's \enquote*{social model of explanation} provides a conceptual bridge between the epistemic pragmatics of \ac{XAI} and the interpretive negotiations that characterize humanistic inquiry.
For art-historical applications---where meaning is constructed through dialogue rather than extracted from data---this social–cognitive model is especially pertinent.
We, thus, combine quantitative and qualitative experiments to examine how explanation methods function not only as diagnostic instruments but as interfaces between human vision and machine inference.

\section{Case Study 1}
\label{sec:study-1}

In the first case study, we evaluate the zero-shot localization capabilities of \ac{CLIP} by examining the afore-introduced \ac{XAI} techniques using large-scale datasets comprising nearly $2{,}000$ art-historical images in total.
In so doing, we determine how effectively each method can identify and delineate objects in domain-specific imagery without any fine-tuning.
The aim of this study is to establish a quantitative foundation for assessing how different explainability methods perform in visually and semantically complex cultural datasets, and to identify which approaches generalize most effectively to art-historical imagery.
Our analysis focuses on comparing these techniques---ScoreCAM, gScoreCAM, GradCAM, GradCAM++, LayerCAM, LeGrad, and CLIP Surgery---under consistent experimental conditions, thereby delineating the empirical contours against which subsequent interpretive analyses (\Cref{sec:study-2}) can be situated.

\begin{figure*}[t!]
    \centering 

    \begin{subfigure}[t]{0.155\textwidth}
        \includegraphics[width=\textwidth]{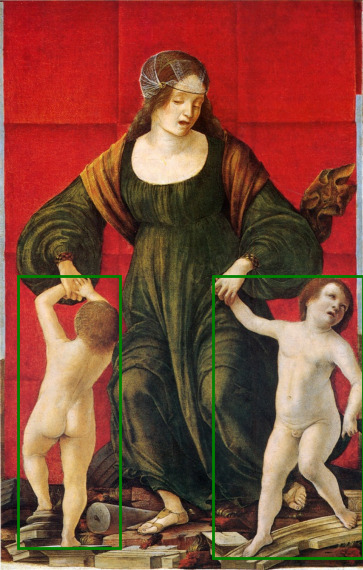}
        \caption{Ground-truth}
    \end{subfigure}%
    \hfill
    \begin{subfigure}[t]{0.155\textwidth}
        \includegraphics[width=\textwidth]{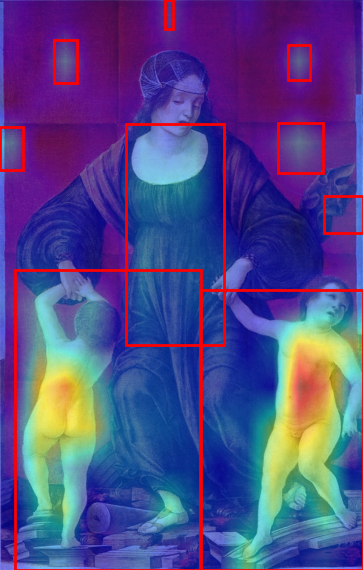}
        \caption{$\tau = 0.20$}
    \end{subfigure}%
    \hfill
    \begin{subfigure}[t]{0.155\textwidth}
        \includegraphics[width=\textwidth]{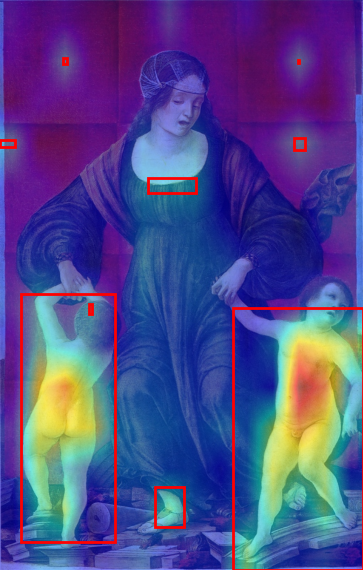}
        \caption{$\tau = 0.30$}
    \end{subfigure}%
    \hfill
    \begin{subfigure}[t]{0.155\textwidth}
        \includegraphics[width=\textwidth]{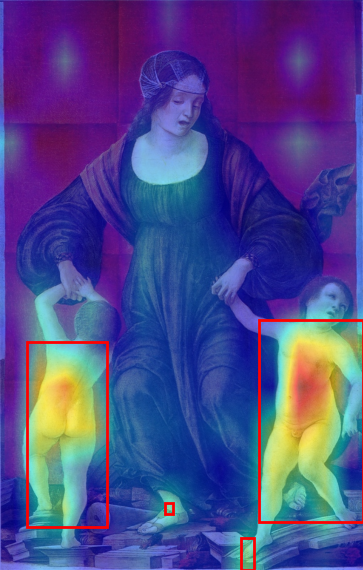}
        \caption{$\tau = 0.40$}
    \end{subfigure}%
    \hfill
    \begin{subfigure}[t]{0.155\textwidth}
        \includegraphics[width=\textwidth]{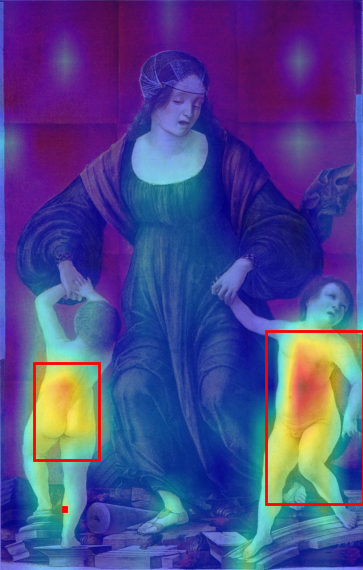}
        \caption{$\tau = 0.50$}
    \end{subfigure}%
    \hfill
    \begin{subfigure}[t]{0.155\textwidth}
        \includegraphics[width=\textwidth]{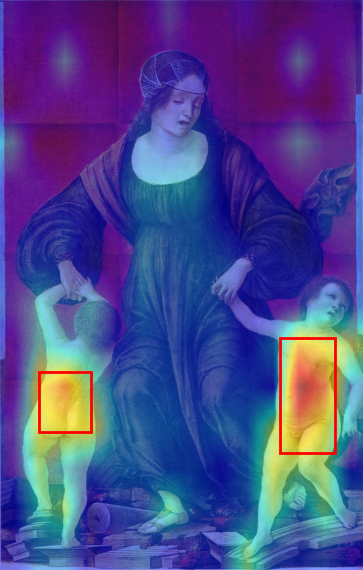}
        \caption{$\tau = 0.60$}
    \end{subfigure}%

    \caption{
        For Ercole de' Roberti's \textit{The Wife of Hasdrubal and Her Children} (c.~1490/1493), the ground-truth bounding boxes for the concept \enquote{nude} are shown in \textit{green} (a).
        The bounding boxes derived from the class-conditional saliency map are displayed at progressively increasing threshold levels $\tau \in \{ 0.20, 0.30, 0.40, 0.50, 0.60 \}$.
    }
    \label{fig:study-1-pipeline}
\end{figure*}

\subsection{Pre-processing Steps}
\label{sec:study-1-preprocessing}

For each method, we generate a class-conditional saliency map, threshold it at $\tau$ to obtain a binary mask, and extract the tightest box around the largest connected component (\Cref{fig:study-1-pipeline}).
These are then compared to ground-truth annotations from the datasets described in \Cref{sec:study-1-dataset}.
While fixed thresholds (e.g., $\tau = 0.2$) can yield plausible maps, they are notoriously sensitive---to the explainability method, the dataset, and even the target class semantics---leading to unstable conclusions.
Following \citet{ChoeOLCAS20}, we therefore adopt a threshold-independent evaluation: for each method, we compute the maximum BoxAcc over $\tau \in [0.2, 0.9]$ via grid search, ensuring that it is evaluated under the most favorable operating conditions.
Unlike \citet{ChoeOLCAS20}, who average BoxAcc across \ac{IoU} thresholds $\delta \in \{ 0.3, 0.5, 0.7 \}$, we report BoxAcc separately at each $\delta$ to identify more granular differences in localization quality.
Formally:
\begin{equation}
    \text{BoxAcc}(\tau, \delta) = \frac{1}{N} \sum_n 1_{\text{IoU} \left( \text{box} \left( s \left( X^{(n)} \right), \tau \right), B^{(n)} \right) \ge \delta}
\end{equation}
where $s \left( X^{(n)} \right)$ is the saliency map for image $X^{(n)}$, $\text{box} \left( s, \tau \right)$ is the tightest bounding box enclosing the largest-area connected component obtained by thresholding $s$ at $\tau$, and $B^{(n)}$ is the corresponding ground-truth bounding box.

All seven methods are evaluated using compatible \ac{CLIP} variants: ScoreCAM, gScoreCAM, GradCAM, GradCAM++, LayerCAM, and CLIP Surgery are applied to the ResNet-50×16 backbone; LeGrad is evaluated on the ViT-B/32 backbone.
For the gradient-based \acp{CAM}, features are extracted from the third ReLU activation in the final bottleneck block of the ResNet \texttt{layer4}, or from the last self-attention heads for \ac{ViT} models.
Gradients of the class score with respect to these activations yield the channel importance weights.
For gradient-free \acp{CAM}, we mask each channel individually by setting all others in the upsampled activation map to zero and compute the cosine similarity between the resulting image embedding and the text prompt to derive the channel importance.
CLIP Surgery applies two inference-time modifications---an adjusted self-attention block and a dual-path feedforward network---to mitigate noisy visualizations without requiring any fine-tuning \citep{clip-surgery}.
All code is implemented in PyTorch and runs on two NVIDIA GeForce RTX 2080 Ti.

\subsection{Data}
\label{sec:study-1-dataset}

As \citet{SchneiderV24} have observed, object-level annotations---especially those specifying spatial localization through bounding boxes---are uncommon in art-historical datasets.
When such annotations do exist, they often deliberately avoid classes representing iconographic concepts, precisely those motifs whose semantic variability resists stable formalization.
The DEArt dataset, for instance, restricts itself to broad descriptive categories such as \enquote{people} or to visually unambiguous but iconographically neutral objects like \enquote{boats} or \enquote{vases} \citep{ReshetnikovML22}.
Yet it is precisely these unstable iconographic categories that most urgently require closer examination if we are to assess whether \acp{VLM} can move beyond generic recognition to something resembling art-historical expertise.

\begin{table}[t!]
\caption{
    Statistics of the \artdl\ \citep{MilaniF21} and \iconart\ \citep{gonthierGLB18} test datasets.
}
\label{tab:study-1-data}

\footnotesize
\begin{tabularx}{\columnwidth}{@{}X*{5}{>{\hsize=0.6\hsize}Y}@{}}
\toprule
    Dataset & \multicolumn{3}{c}{Images} & Boxes & Classes\\
\cmidrule(lr){2-4}
    & Total & Positive & Negative & & \\
\midrule
    \artdl   & 4,166 &  808  & 3,358 & 3,793 & 59 \\
    \iconart & 1,480 & 1,037 &   443 & 4,931 & 10 \\
\bottomrule
\end{tabularx}
\end{table}
\begin{table*}[t!]
\caption{
    Bounding box detection results are reported for the IconArt \citep{gonthierGLB18} and ArtDL \citep{MilaniF21} test sets, using the highest BoxAcc obtained across all binarization thresholds $\tau$.
    The best performing approach per test set is indicated in bold.
}
\label{tab:study-1-results}

\footnotesize
\begin{tabularx}{\textwidth}{@{}X l RRRRRRRR@{}}
    \toprule
    Dataset & Method & \multicolumn{4}{c}{IoU $\ge0.30$} & \multicolumn{4}{c}{IoU $\ge0.50$} \\
    \cmidrule(lr){3-6}
    \cmidrule(l){7-10}
    & & {BoxAcc} & $\text{BoxAcc}_S$ & $\text{BoxAcc}_M$ & $\text{BoxAcc}_L$ & {BoxAcc} & $\text{BoxAcc}_S$ & $\text{BoxAcc}_M$ & $\text{BoxAcc}_L$ \\
    \midrule
    \iconart   
        & CLIP Surgery & \bfseries 0.2876 & \bfseries 0.0862 & 0.2447 & \bfseries 0.6623 & \bfseries 0.1482 & 0.0219 & 0.0807 & \bfseries 0.4070 \\
        & LeGrad       & 0.2722 & 0.0849 & \bfseries 0.2585 & 0.6436 & 0.1369 & \bfseries 0.0232 & \bfseries 0.0981 & 0.3808 \\
        & ScoreCAM     & 0.2411 & 0.0433 & 0.2264 & 0.6230 & 0.1040 & 0.0107 & 0.0852 & 0.2953 \\
        & gScoreCAM    & 0.2344 & 0.0679 & 0.2356 & 0.6024 & 0.1121 & 0.0174 & 0.0843 & 0.3208 \\
        & GradCAM      & 0.1391 & 0.0666 & 0.2145 & 0.2509 & 0.0355 & 0.0156 & 0.0660 & 0.0624 \\
        & GradCAM++    & 0.1584 & 0.0277 & 0.0880 & 0.4432 & 0.0584 & 0.0067 & 0.0192 & 0.1779 \\
        & LayerCAM     & 0.1783 & 0.0420 & 0.1769 & 0.4694 & 0.0627 & 0.0094 & 0.0577 & 0.1823 \\
    \midrule
    \artdl
        & CLIP Surgery & \bfseries 0.5228 & \bfseries 0.2069 & \bfseries 0.4946 & \bfseries 0.7487 & \bfseries 0.3019 & \bfseries 0.0722 & \bfseries 0.2205 & \bfseries 0.5297 \\
        & LeGrad       & 0.4382 & 0.1778 & 0.4280 & 0.7035 & 0.2552 & 0.0458 & 0.1868 & 0.4680 \\
        & ScoreCAM     & 0.3557 & 0.0667 & 0.3055 & 0.6457 & 0.1672 & 0.0125 & 0.1087 & 0.3441 \\
        & gScoreCAM    & 0.3815 & 0.0889 & 0.3675 & 0.6627 & 0.1727 & 0.0208 & 0.1378 & 0.3497 \\
        & GradCAM      & 0.2684 & 0.1056 & 0.3851 & 0.2790 & 0.0701 & 0.0278 & 0.0988 & 0.0787 \\
        & GradCAM++    & 0.2418 & 0.0403 & 0.1493 & 0.4890 & 0.1168 & 0.0069 & 0.0237 & 0.2484 \\
        & LayerCAM     & 0.2476 & 0.0583 & 0.2052 & 0.4578 & 0.0873 & 0.0097 & 0.0582 & 0.1590 \\
    \bottomrule
\end{tabularx}
\end{table*}

For this reason, our evaluation focuses on two datasets that explicitly provide iconographic content: \iconart\ \citep{gonthierGLB18} and \artdl\ \citep{MilaniF21} (\cref{tab:study-1-data}).
Both provide annotations not only for figures (e.g., \enquote{Saint Sebastian}) but also for attributes and symbols (e.g., \enquote{Ointment Jar}).
Of these, \artdl\ is the broader in scope, comprising $10$ saints and $49$ attributes.
However, the usefulness of these datasets as benchmarks is limited by their distribution: both have pronounced long-tails.
In \iconart, three generic categories---\enquote{beard} ($21.98$\,\%), \enquote{angel} ($21.15$\,\%), and \enquote{nudity} ($15.39$\,\%)---account for nearly two-thirds of all annotations, while classes of greater art-historical specificity such as \enquote{Saint Sebastian} ($1.66$\,\%) or \enquote{Crucifixion of Jesus} ($2.21$\,\%) appear only rarely.
\artdl\ displays the same dynamic even more strongly: the class \enquote{face} ($21.28$\,\%) dominates more semantically charged categories, followed by \enquote{Mary} ($7.46$\,\%) and \enquote{Baby Jesus} ($6.04$\,\%).
Attributes critical for saint identification---such as the \enquote{lily} ($0.92$\,\%)---are relegated to statistical noise, with over half of the classes represented by fewer than $50$ instances.
The implications are methodological as much as statistical.
Evaluating these corpora has the potential to reward models for recognizing abundant, generic classes while obscuring failures in categories of genuine iconographic expertise.
In other words, a model may perform convincingly on aggregate measures while exhibiting no real grasp of the iconographic logics central to art-historical interpretation.
However, both \iconart\ and \artdl\ are indispensable precisely because they are the only datasets that even begin to capture iconographic detail.
They are best understood, then, not as definitive benchmarks but as starting points---\enquote*{partial ground truths,} so to speak, that expose, as much as they enable, the epistemic challenges of employing \acp{VLM} in art history.

\subsection{Results}
\label{sec:study-1-results}

\Cref{tab:study-1-results} reports the comparative performance of all evaluated methods.
Across both the \iconart\ and \artdl\ test sets, CLIP Surgery achieves higher accuracy scores than all other methods, particularly at the more permissive \ac{IoU} threshold of $0.30$.
On the \artdl\ test set, it obtains a BoxAcc of $52.28$\,\% at an \ac{IoU} threshold of $0.30$---an absolute improvement of almost $9$ points over the second-best method, LeGrad ($43.82$\,\%); this is also evident at the stricter \ac{IoU} threshold of $0.50$, where CLIP Surgery yields a BoxAcc of $30.19$\,\%.
This advantage extends across object scales: at \ac{IoU} $\ge0.30$, CLIP Surgery demonstrates the highest accuracy for small ($20.69$\,\%), medium ($49.46$\,\%), and large objects ($74.87$\,\%).
We observe some exceptions at the class level---for instance, \enquote{baby Jesus} performs better with LeGrad ($76.79$\,\%) than with CLIP Surgery ($48.21$\,\%)---but these are relatively isolated cases.
However, for some classes LeGrad is preferable: for instance, \enquote{baby Jesus} achieves $76.79$\,\% with LeGrad but only $48.21$\,\% with CLIP Surgery.
Even when the threshold rises to \ac{IoU} $\ge 0.50$, where precise localization is more challenging, it remains the most accurate in all size categories.
In contrast, gradient-based methods---GradCAM, GradCAM++, and LayerCAM---experience significant performance degradation under both \ac{IoU} thresholds.
On the \iconart\ test set, CLIP Surgery similarly achieves the highest BoxAcc at \ac{IoU} $\ge 0.30$ ($28.76$\,\%); yet LeGrad slightly outperforms it in medium-object accuracy ($25.85$\,\% versus $24.47$\,\%).
In addition, object-level analysis reveals considerable differences between the two leading methods.
For instance, in \iconart, CLIP Surgery recognizes the class \enquote{Mary} with an accuracy of $85.82$\,\% for large objects, compared to $76.62$\,\% with LeGrad.
The difference is even greater for \enquote{nudity,} where CLIP Surgery attains $66.16$\,\% versus LeGrad's $50.25$\,\%.
At the stricter \ac{IoU} $\ge 0.50$, CLIP Surgery reclaims overall superiority (with a BoxAcc of $14.82$\,\%), with LeGrad performing marginally better on small and medium objects.
Gradient-based methods perform still more poorly here, underscoring their limited transferability to iconographic material.

The lower overall accuracy observed on \iconart\ relative to \artdl\ arises from both structural and semantic differences between the two corpora.
First, \iconart\ contains a higher proportion of small objects---those occupying $\leq 1$\,\% of the image area---with $46.30$\,\% of all instances, versus $18.98$\,\% in \artdl.
Small objects are inherently more challenging to detect and classify, because they provide fewer pixels for feature extraction and are more susceptible to background clutter; the overrepresentation of small objects consequently reduces the aggregate performance on \iconart.
Second, the datasets have different epistemic scopes.
\artdl\ is broader and more generic, including categories that can be recognized without specialized iconographic knowledge, e.g., \enquote{beard.}
\iconart, by contrast, focuses on a small number of historically charged motifs---such as the \enquote{Crucifixion of Jesus}---whose correct identification depends on contextual and narrative cues rather than isolated attributes.
These scenes are formally and semantically dense: their complexity and entanglement with related subthemes (e.g., episodes from Christ's Passion) expose the limitations of models optimized for visual generality.

\begin{figure*}[t!]
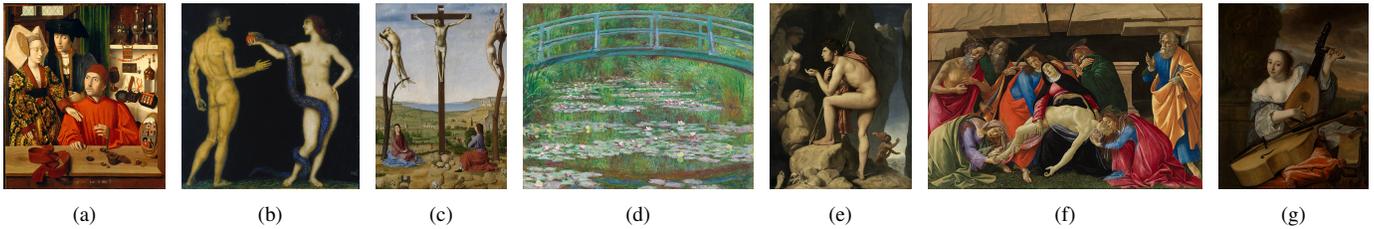

  \centering 

  \foreach \img [count=\i] in {
    a-goldsmith-in-his-shop.jpg,
    adam-and-eve.jpg,
    calvary.jpg,
    japanese-footbridge.jpg,
    oedipus-and-the-sphinx.jpg,
    the-lamentation.jpg,
    the-musician.jpg
  }{%
    \imagemap[2.46cm]{images/\img}%
    \ifnum\i<7 \hfill\fi
  }

  \caption{
    Seven artworks were selected for the online study, each paired with two target classes:
    Petrus Christus, \textit{A Goldsmith in his Shop} (1449; a) with \enquote{convex mirror} and \enquote{girdle};
    Franz von Stuck, \textit{Adam and Eve} (c.~1920; b) with \enquote{arm outstretched} and \enquote{snake};
    Antonello da Messina, \textit{Calvary} (1475; c) with \enquote{John} and \enquote{thief};
    Claude Monet, \textit{Japanese Footbridge} (1899; d) with \enquote{bridge} and \enquote{flower};
    Jean-Auguste-Dominique Ingres, \textit{Oedipus and the Sphinx} (1808; e) with \enquote{left foot} and \enquote{Sphinx};
    Sandro Botticelli, \textit{The Lamentation} (c.~1490; f) with \enquote{sword} and \enquote{Virgin Mary};
    Bartholomeus van der Helst, \textit{The Musician} (1662; g) with \enquote{lustful} and \enquote{sheet music.}
  }
  \label{fig:study-2-data}
\end{figure*}

\section{Case Study 2}
\label{sec:study-2}

So, what can we establish at this point?
Across two large-scale datasets, our evaluation shows that CLIP Surgery consistently outperforms all other methods, with LeGrad emerging---almost unequivocally---as the second-best approach.
Yet the discrepancies between the two datasets, \iconart\ and \artdl, are instructive.
While both annotate iconographic figures and attributes, they, inevitably, only partially reflect the broader iconographic heterogeneity that defines art-historical material.
Our large-scale approach therefore provides valuable, albeit ultimately limited, perspectives as the selected classes naturally constrain the horizon of possible inquiry: we can assess performance within these bounds, but we cannot assume that the epistemic space of art history is adequately represented.\footnote{This limitation, of course, applies to all machine-vision research, since available datasets---shaped by resource constraints---necessarily offer a highly selective view of historical material tailored to specific research questions.}
The second case study thus deliberately shifts focus.
It addresses the gap already identified by \citet{Miller19}, namely the absence of human-centered evaluations of \ac{XAI} methods.
While the first case study focused on measuring the accuracy of these methods and assigning numerical performance scores, this second case study explores interpretability---asking not only \textit{what} these models predict, but \textit{how} their explanations are understood by human users.
To approximate the practical variability of art-historical research---resistant as it is to fixed classification---we employ a broader and more diverse selection of images and categories.
Our methodological scope remains consistent with the first study, encompassing the same explanation methods---ScoreCAM, gScoreCAM, GradCAM, GradCAM++, LayerCAM, LeGrad, and CLIP Surgery.
But the focus is no longer on algorithmic performance.
Rather, it tests whether these methods can make visible the complex visual logics that art history engages with, and whether they succeed, or fail, in approximating the visual concepts that constitute the field.

\subsection{Experimental Design}
\label{sec:study-2-method}

To assess the degree to which algorithm-generated saliency maps align with human perceptions of visual importance, we conducted a within-subjects online study using \textit{SoSci Survey} between June and July 2025.\footnote{\href{https://www.soscisurvey.de/} (accessed on \today).}
After reviewing and accepting an informed consent form, participants provided socio-demographic information, including age, gender identity, education level, and professional status.
On each subsequent page, they were shown one of seven artworks (\Cref{fig:study-2-data}) and asked to use their mouse to annotate regions they deemed relevant to a specified class---e.g., \enquote{convex mirror} or \enquote{Virgin Mary}.
The selection of artworks spanned a wide chronological and stylistic range, from Renaissance devotional works to early twentieth-century Symbolism.
This diversity was intentional: it ensured that annotators were exposed to heterogeneous visual traditions and compositional strategies.
Some target classes referred to discrete, visually localized elements (e.g., \enquote{bridge}), while others invoked symbolic or abstract categories (e.g., \enquote{lustful}).
The participants then viewed seven saliency maps for each image-class pair, again generated by the following explanation methods: ScoreCAM, gScoreCAM, GradCAM, GradCAM++, LayerCAM, LeGrad, and CLIP Surgery.
They were asked to order these heatmaps according to how well they reflected the regions they had previously identified as important.
This ranking served as a subjective measure of the alignment between human visual attention and the algorithmic saliency outputs.
To minimize order effects, both the sequence of image–class pairs and the presentation order of saliency maps were independently randomized for each participant.
Only those participants who completed at least four of the $14$ ranking tasks were included in the final analysis.

\subsection{Participants}
\label{sec:study-2-dataset}

Participants were recruited from the student populations of the University of Munich and the University of Göttingen;
participation was voluntary and not incentivised.
The study involved $33$ participants, of whom $21.21$\,\% identified as male, $75.76$\,\% as female, and $3.03$\,\% as diverse.
Participant ages ranged from $18$ to over $65$ years ($\text{M} = 42.21$; $\text{SD} = 19.08$).
Regarding educational background, $39.39$\,\% held qualifications for university entry, while $45.46$\,\% had completed a degree at either a university or a university of applied sciences.
Students made up the largest subgroup, accounting for $54.54$\,\% of all respondents.
In terms of art-historical expertise, $62.50$\,\% reported basic knowledge (e.g., gained through introductory coursework), $21.88$\,\% reported intermediate proficiency (gained through a bachelor's degree or initial professional experience), and the remainder described themselves as advanced or expert.
The demographic profile thus reflects a fairly diverse sample in terms of both age and educational attainment.
As a complete-case analysis would have required excluding $37.93$\,\% of participants---had full rankings across all seven methods and images been required---or $8.86$\,\% of the individual rankings---if all methods had to be ranked per image---we chose to impute missing rank positions using the \ac{MICE} algorithm over $20$ iterations \citep{mice}.
Although the sample size of $n = 33$ is at the lower end of recommended thresholds for within-subjects tests, power analysis indicates sufficient sensitivity to detect effects of medium magnitude (Kendall's $W \geq 0.30$) with an estimated probability of $\approx 80$--$85$\,\%.
For the purposes of this investigation, such values are deemed adequate: the study is not intended to establish statistically significant differences between these specific artworks, but rather to assess---at least to some extent---the general ranking of saliency maps in terms of their alignment with human annotations.
The central objective is to determine whether a modest, curated sample obtained through a user study can already reveal broader patterns that are informative for evaluating the applicability of \ac{XAI} methods in art-historical contexts.

\begin{figure*}[!t]
    \centering
    \includegraphics[width=0.92\textwidth]{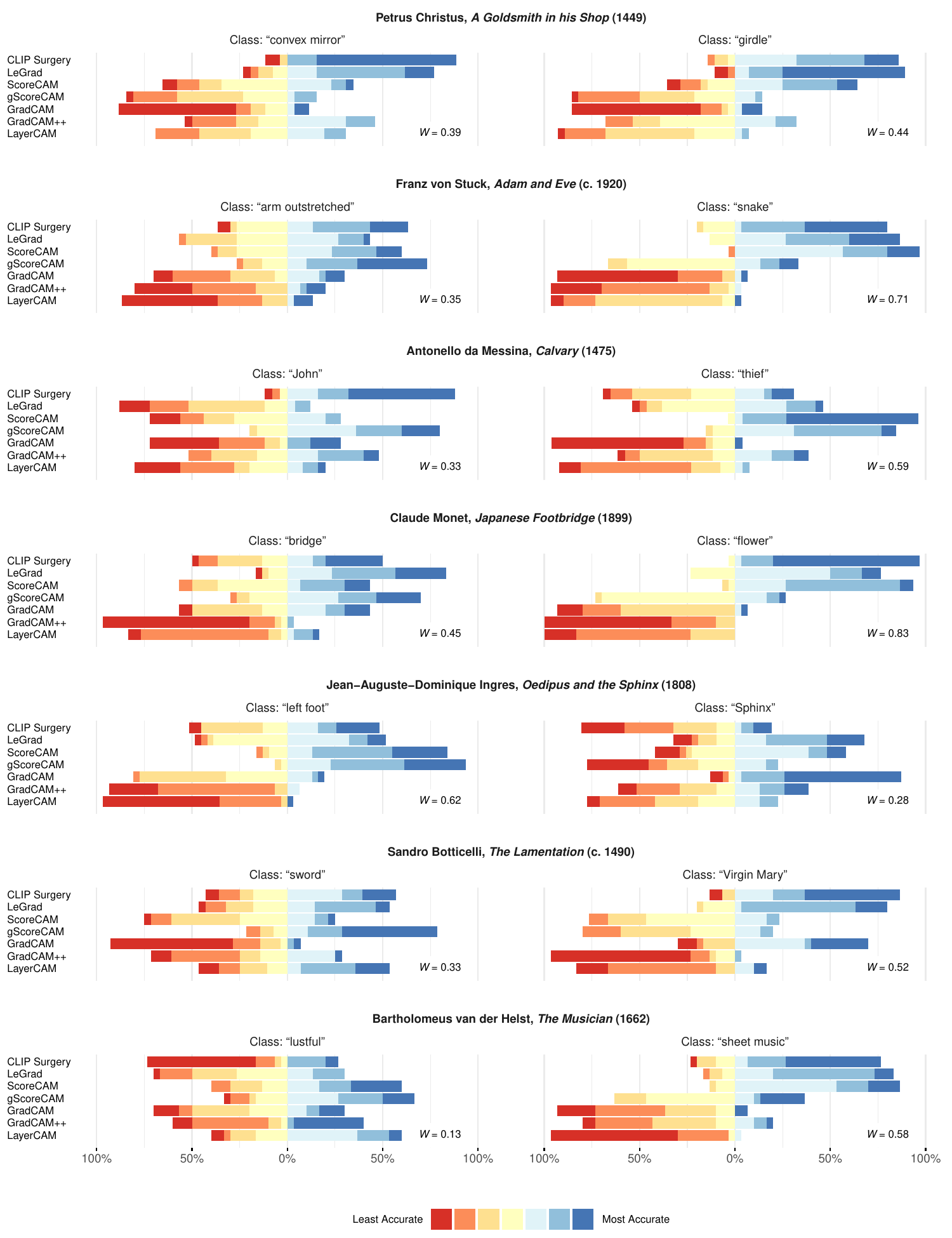}

    \caption{
        Evaluation results are shown as divergent stacked bar charts, comparing seven visual explainability methods for each image and class.
        Colors range from \textit{red} (\enquote{least accurate}) to \textit{blue} (\enquote{most accurate}).
        Kendall's $W$ is reported to assess inter-rater reliability.
    }
    \label{fig:study-2-ranking-results}
\end{figure*}

\subsection{Results}
\label{sec:study-2-results}

\Cref{fig:study-2-ranking-results} shows divergent stacked bar charts of ranking distributions for each image-class pair.
The charts illustrate the extent to which participants agreed or disagreed in their assessments, making it easier to identify patterns of consensus or divergence across images.
To formally quantify inter-rater agreement, Kendall’s $W$ is reported for each pair: higher values reflect stronger reliability, while lower values indicate greater variability in judgments.
Across the images, three techniques---CLIP Surgery, LeGrad, and ScoreCAM---achieve the highest mean rank positions, indicating that participants perceived these methods as most faithfully highlighting their annotated regions; gScoreCAM also performs strongly, but shows slightly greater variability. 
In contrast, GradCAM, GradCAM++, and LayerCAM are consistently ranked towards the bottom, suggesting that their gradient-based heatmaps align less closely with the participants' own saliency judgments.
These patterns can be identified regardless of participants' art-historical expertise.
While CLIP Surgery is favored by participants with basic knowledge, those with at least intermediate proficiency have a slight preference for LeGrad; at this level, gScoreCAM also approaches the top-performing group.
Beyond these minor differences, no further---and in particular, no statistically significant---effects are observed.
For visually well-defined or spatially localized targets---e.g., the \enquote{snake} in von Stuck's \textit{Adam and Eve} (\Cref{fig:study-2-annotation-results-adam-and-eve}) or the \enquote{left foot} in Ingres's \textit{Oedipus and the Sphinx} (\Cref{fig:study-2-annotation-results-oedipus-and-the-sphinx})---participants' rankings converge tightly, with Kendall's $W$ indicating strong inter-rater reliability ($W = 0.71$ and $W = 0.62$, respectively).
A similar pattern emerges for the flowers beneath the bridge in Monet's painting ($W = 0.83$), pointing to a shared perception of salient features (\Cref{fig:study-2-annotation-results-japanese-footbridge}).
By contrast, more diffuse or interpretative categories, such as \enquote{lustful} (\Cref{fig:study-2-annotation-results-the-musician}) or the \enquote{Sphinx} (\Cref{fig:study-2-annotation-results-oedipus-and-the-sphinx}), yield widely dispersed rankings and low Kendall's $W$, with no single method emerging as dominant; this reflects the intrinsic ambiguity of these  higher-order visual concepts.
However, the difficulty of some classes also becomes evident in cases where even human annotators struggle to establish consistent associations.
Take, for example, Ingres's \textit{Oedipus and the Sphinx}, in which three distinct \enquote{left feet} would, in principle, need to be annotated: (1)~the left foot of Oedipus, firmly bracing his weight; (2)~the left foot of the Sphinx, obscured in shadow and therefore difficult to recognize---mirrored by the sparse human annotations in \Cref{fig:study-2-annotation-results-oedipus-and-the-sphinx}, and by the failure of saliency maps to capture it; and (3)~a severed left foot at the lower edge of the canvas, possibly belonging to one of the Sphinx's earlier victims.
In this case, annotators likely registered only the most prominent instance, suggesting insufficient time for close observation.
Yet in other examples, the challenge arises less from attention than from limited knowledge of the relevant visual structures associated with a given class.
An interesting case is the red wedding girdle in Petrus Christus's \textit{Goldsmith}, which projects into the viewer's space over the shop's ledge, yet is far less frequently annotated than the woman's hip belt (\Cref{fig:study-2-annotation-results-a-goldsmith-in-his-shop}).
The issue is even more evident in Botticelli's \textit{Lamentation}.
While the dead Christ rests in the Virgin Mary's lap, two additional figures---Mary Magdalene and, very likely, Mary of Clopas---gently support his head and feet.
Yet participants frequently mislabeled these figures as the Virgin Mary as well (\Cref{fig:study-2-annotation-results-the-lamentation}).

\section{Discussion}
\label{sec:discussion}

The case studies reveal two interrelated factors.
(1)~\textit{Ambiguity of concepts}.
In art-historical imagery, target classes often are interpretive rather than fixed, and thus resist stable localization.
In Botticelli's \textit{Lamentation}, the three Marys mourning over Christ are visually similar, so that non-specialist annotators might confuse them (\Cref{fig:study-2-annotation-results-the-lamentation}).
Here, \textit{discernibility} emerges as a criterion: where classes are both concrete and spatially bounded, annotators converge; where they are symbolic, context-dependent, or reliant on art-historical expertise, judgments diverge and inter-rater reliability declines.
Ground-truth annotations, in this sense, are never exhaustive.
Methods that highlight only the most visually dominant instance of a class therefore cannot easily be dismissed as inadequate, since they nonetheless establish a valid link between text and image.
Yet ambiguity does not arise only from annotators: it is also encoded in the model itself.
This leads to:
(2)~\textit{Limits of representation}.
Saliency methods cannot recover what the model itself fails to encode.
If a concept does not appear as a localized hotspot in \ac{CLIP}'s latent space, attribution will necessarily remain diffuse, regardless of the post-hoc technique employed.
This is evident in Antonello da Messina's \textit{Calvery}: the thieves flanking Christ are inconsistently mapped---ScoreCAM and gScoreCAM weakly associate both figures with the prompt \enquote{thief,} while LeGrad isolates only the feet of the right-hand figure (\Cref{fig:study-2-annotation-results-calvary}).
Such incoherence suggests that \ac{CLIP} does not encode \enquote{thief} as a transferable visual concept.
The difficulty may partly stem from the unnatural posture of the figures, but---more fundamentally---it reflects the fragmentary nature of the training corpus.
In contrast to the photographic imagery that dominates \ac{CLIP}'s dataset, crucifixion scenes are relatively uncommon, with peripheral figures such as the thieves being especially under-specified.
The challenge is further complicated by the semantic breadth of the term \enquote{thief.}
Unlike \enquote{Christ} or \enquote{cross,} which correspond to highly codified and visually stable iconographic forms, \enquote{thief} has no fixed template: it may denote an anonymous criminal, a masked burglar, or---in the Passion narrative---the unnamed \enquote*{good} and \enquote*{bad} thieves, who are distinguished only by their position relative to Christ and, in some traditions, by subtle gestures or expressions.
Ambiguity, in this case, is not simply a problem of human perception, but a structural property of the machine's representational logic.
Yet across both cases, relative performance of gradient-based, score-based, and hybrid methods remains largely stable.
This indicates that small-scale, human-centered studies can support robust comparative evaluation, while large-scale evaluation on pre-existing datasets can extend such results---though only for a narrow subset of art-historically relevant categories.
More importantly, these studies also function diagnostically: they reveal not only how well attribution methods capture model activations, but also how models themselves mediate art-historical concepts.
Human studies, carefully designed, thus can produce findings that generalize to broader validations while simultaneously exposing the cultural and epistemic imaginaries embedded in machine vision.

For real-time or ad hoc explanation needs, additional considerations are required.
ScoreCAM obtains channel weights by computing the forward-pass score for each activation map---i.e. it performs one forward pass per channel (i.e., $C$ forward passes per image)---to generate a single heatmap (e.g., $3{,}072$ forward passes for RN50×16), which makes it slow for on-the-fly explanations \citep{scorecam}.
gScoreCAM alleviates this bottleneck by selecting only the top-$k$ channels (typically $k=300$), reducing the number of forward passes to about $0.1C \approx 307$ for $C = 3{,}072$ \citep{gscorecam}.
By contrast, gradient-based methods (GradCAM, GradCAM++, LayerCAM, and variants such as LeGrad) require one forward pass and one backward pass to compute \acp{CAM}.
However, for real-time or ad hoc explanations, CLIP Surgery is particularly advantageous, as it requires only a single modified forward pass without any gradient computations or backpropagation \citep{clip-surgery}.\footnote{Actual runtime performance depends on hardware and implementation details. We recommend empirical benchmarking of all methods on the target system to obtain accurate latency estimates before deployment.}
Beyond computational efficiency, these methodological differences also influence the epistemic character of the resulting explanations.
Multi-pass methods like ScoreCAM and gScoreCAM often produce smoother, less noisy saliency maps, but their cost makes them impractical for interactive settings.
Gradient-based methods, while faster, can suffer from gradient saturation or overemphasis on high-level features, resulting in instability or sensitivity across inputs.
CLIP Surgery, though extremely efficient, forgoes gradient information entirely; its outputs should thus be interpreted as approximations of model \enquote*{attention} rather than as exhaustive mappings of activation contributions.
In art-historical contexts, this distinction is nontrivial: when interpreting ambiguous or symbolical imagery, the choice of method not only constrains computational performance but also shapes the interpretive claims about what the model \enquote*{perceives.}
Real-time explanation pipelines, then, must balance latency constraints with the need for stable, interpretable visualizations---accepting that faster methods may favor responsiveness and accessibility, while slower methods may deliver higher resolution or fidelity but at the cost of usability in exploratory settings.

\section{Conclusions}
\label{sec:conclusions}

Our case studies---addressing the questions raised in \Cref{sec:introduction}---demonstrate that the epistemic promise of \ac{XAI} in digital art history is both methodological and hermeneutic.
With regard to the first question (\textit{How effectively do \ac{XAI} methods localize iconographic objects in artworks under zero-shot conditions without fine-tuning?}), our quantitative evaluation shows that CLIP Surgery, which is specifically adapted to \ac{CLIP}'s dual-encoder architecture, consistently outperforms general-purpose gradient- or score-based methods.
Even when trained primarily on non-art-historical imagery, CLIP Surgery can delineate a broad range of iconographic concepts, particularly when the target classes are visually distinct or spatially constrained; its accuracy, however, declines for semantically complex motifs.
The limitation here lies not in the \ac{XAI} method itself but in the representational granularity of the embedding space: \ac{CLIP} does not, in fact, \enquote*{see} an object in its historicity, but only the statistical residue of an already mediated image-world.
Turning to the second question (\textit{Does the visual relevance of these maps correspond to human judgments?}), our user study partially confirms this observation.
Participants generally preferred CLIP Surgery, LeGrad, and ScoreCAM, whose saliency maps closely approximated human annotations.
Yet for abstract or context-dependent classes, agreement and, thus, inter-rater reliability decreased.
This divergence underscores a central tension: while saliency maps can reproduce certain aspects of perceptual attention, they cannot replicate the interpretive depth of the art-historical gaze.
The third question (\textit{Which factors, such as object size and concept abstraction, drive performance differences?}) reveals both structural and semantic determinants.
Localization accuracy correlates strongly with object size---and, thus, visual prominence---but it is equally influenced by conceptual stability.
Classes with more consistent visual referents (e.g., \enquote{bridge} or \enquote{snake}) yielded higher accuracy scores than those with diffuse, symbolic, or art-historically specific meanings (e.g., \enquote{lustful} or \enquote{Virgin Mary}).
These findings suggest that model performance reflects how, and to what extent, a concept exists within the model's learned visual ontology.
Returning to a broader question raised in \Cref{sec:introduction}---whether \ac{XAI} discloses a model's internal conceptual structure or merely aestheticizes its opacity---our answer must be: it depends.
The visual legibility of a saliency map can be deceptive as it does not imply epistemic transparency.
Such maps expose the internal dynamics of how certain features or tokens activate within an embedding geometry, while concealing the historical, cultural, and linguistic priors that render those activations meaningful.
What is visualized, therefore, is not the model's \enquote*{understanding} of an artwork but the projection of human interpretive desire onto computational artifacts that can only approximate meaning.
Explainability in digital art history must, in this context, be conceived as a dialogical process between human and machine vision, as elaborated by \citet{Miller19}---demanding critical awareness of the epistemic imaginaries through which models \enquote*{see.}
\ac{XAI} outputs should therefore be read not as self-sufficient explanations but as prompts for further hermeneutic inquiry.

\section*{Acknowledgements}

This work was funded in part by the German Research Foundation (Deutsche Forschungsgemeinschaft; DFG) under project number 510048106.
We thank Hubertus Kohle, Ralph Ewerth, Eric Müller-Budack, Matthias Springstein, and Julian Stalter for their insightful discussions and helpful comments on the subject matter.

\printbibliography

@inproceedings{gonthierGLB18,
  author       = {Nicolas Gonthier and
                  Yann Gousseau and
                  Said Ladjal and
                  Olivier Bonfait},
  editor       = {Laura Leal{-}Taix{\'{e}} and
                  Stefan Roth},
  title        = {Weakly Supervised Object Detection in Artworks},
  booktitle    = {Computer Vision -- {ECCV} 2018 Workshops},
  series       = {Lecture Notes in Computer Science},
  volume       = {11130},
  pages        = {692--709},
  publisher    = {Springer},
  year         = {2018},
  doi          = {10.1007/978-3-030-11012-3\_53},
}

@inproceedings{ReshetnikovML22,
  author       = {Artem Reshetnikov and
                  Maria{-}Cristina V. Marinescu and
                  Joaquim Mor{\'{e}} L{\'{o}}pez},
  editor       = {Leonid Karlinsky and
                  Tomer Michaeli and
                  Ko Nishino},
  title        = {{DEArt}: Dataset of European Art},
  booktitle    = {Computer Vision -- {ECCV} 2022 Workshops},
  series       = {Lecture Notes in Computer Science},
  volume       = {13801},
  pages        = {218--233},
  publisher    = {Springer},
  year         = {2022},
  doi          = {10.1007/978-3-031-25056-9\_15},
}

@article{SchneiderV24,
  author       = {Stefanie Schneider and
                  Ricarda Vollmer},
  title        = {Poses of People in Art: {A} Dataset for Human Pose Estimation in Digital Art History},
  journal      = {{ACM} Journal on Computing and Cultural Heritage},
  volume       = {17},
  number       = {4},
  pages        = {61:1--61:19},
  year         = {2024},
  doi          = {10.1145/3696455},
}

@article{MilaniF21,
  author       = {Federico Milani and
                  Piero Fraternali},
  title        = {A Dataset and a Convolutional Model for Iconography Classification
                  in Paintings},
  journal      = {{ACM} Journal on Computing and Cultural Heritage},
  volume       = {14},
  number       = {4},
  pages        = {46:1--46:18},
  year         = {2021},
  doi          = {10.1145/3458885},
}

@article{Miller19,
  author       = {Tim Miller},
  title        = {Explanation in Artificial Intelligence: Insights from the Social Sciences},
  journal      = {Artificial Intelligence},
  volume       = {267},
  pages        = {1--38},
  year         = {2019},
  doi          = {10.1016/J.ARTINT.2018.07.007},
}

@inproceedings{gscorecam,
  author       = {Peijie Chen and
                  Qi Li and
                  Saad Biaz and
                  Trung Bui and
                  Anh Nguyen},
  editor       = {Lei Wang and
                  Juergen Gall and
                  Tat{-}Jun Chin and
                  Imari Sato and
                  Rama Chellappa},
  title        = {gScoreCAM: What Objects Is {CLIP} Looking At?},
  booktitle    = {Computer Vision -- {ACCV} 2022 -- 16th Asian Conference on Computer Vision},
  series       = {Lecture Notes in Computer Science},
  volume       = {13844},
  pages        = {588--604},
  publisher    = {Springer},
  year         = {2022},
  doi          = {10.1007/978-3-031-26316-3\_35},
}

@article{clip-surgery,
  author       = {Yi Li and
                  Hualiang Wang and
                  Yiqun Duan and
                  Jiheng Zhang and
                  Xiaomeng Li},
  title        = {A Closer Look at the Explainability of Contrastive Language-image Pre-training},
  journal      = {Pattern Recognition},
  volume       = {162},
  pages        = {111409},
  year         = {2025},
  doi          = {10.1016/J.PATCOG.2025.111409},
}

@inproceedings{scorecam,
  author       = {Haofan Wang and
                  Zifan Wang and
                  Mengnan Du and
                  Fan Yang and
                  Zijian Zhang and
                  Sirui Ding and
                  Piotr Mardziel and
                  Xia Hu},
  title        = {{Score-CAM:} Score-Weighted Visual Explanations for Convolutional Neural Networks},
  booktitle    = {2020 {IEEE/CVF} Conference on Computer Vision and Pattern Recognition, {CVPR} Workshops 2020},
  pages        = {111--119},
  publisher    = {Computer Vision Foundation / {IEEE}},
  year         = {2020},
  doi          = {10.1109/CVPRW50498.2020.00020},
}

@inproceedings{gradcam,
  author       = {Ramprasaath R. Selvaraju and
                  Michael Cogswell and
                  Abhishek Das and
                  Ramakrishna Vedantam and
                  Devi Parikh and
                  Dhruv Batra},
  title        = {{Grad-CAM:} Visual Explanations from Deep Networks via Gradient-Based Localization},
  booktitle    = {{IEEE} International Conference on Computer Vision, {ICCV} 2017},
  pages        = {618--626},
  publisher    = {{IEEE} Computer Society},
  year         = {2017},
  doi          = {10.1109/ICCV.2017.74},
}

@inproceedings{gradcam++,
  author       = {Aditya Chattopadhyay and
                  Anirban Sarkar and
                  Prantik Howlader and
                  Vineeth N. Balasubramanian},
  title        = {{Grad-CAM++:} Generalized Gradient-Based Visual Explanations for Deep Convolutional Networks},
  booktitle    = {2018 {IEEE} Winter Conference on Applications of Computer Vision, {WACV} 2018},
  pages        = {839--847},
  publisher    = {{IEEE} Computer Society},
  year         = {2018},
  doi          = {10.1109/WACV.2018.00097},
}

@article{layercam,
  author       = {Peng{-}Tao Jiang and
                  Chang{-}Bin Zhang and
                  Qibin Hou and
                  Ming{-}Ming Cheng and
                  Yunchao Wei},
  title        = {{LayerCAM:} Exploring Hierarchical Class Activation Maps for Localization},
  journal      = {{IEEE} Transactions on Image Processing},
  volume       = {30},
  pages        = {5875--5888},
  year         = {2021},
  doi          = {10.1109/TIP.2021.3089943},
}

@misc{legrad,
  author       = {Walid Bousselham and
                  Angie W. Boggust and
                  Sofian Chaybouti and
                  Hendrik Strobelt and
                  Hilde Kuehne},
  title        = {{LeGrad:} An Explainability Method for Vision Transformers via Feature Formation Sensitivity},
  year         = {2024},
  eprinttype   = {arXiv},
  eprint       = {2404.03214},
}

@article{mice,
  author       = {S. van Buuren and
                  K. Groothuis-Oudshoorn},
  title        = {{mice}: Multivariate Imputation by Chained Equations in {R}},
  journal      = {Journal of Statistical Software},
  volume       = {45},
  number       = {3},
  pages        = {1--67},
  year         = {2011},
  doi          = {10.18637/jss.v045.i03},
}

@inproceedings{BhallaOSCL24,
  author       = {Usha Bhalla and
                  Alex Oesterling and
                  Suraj Srinivas and
                  Fl{\'{a}}vio P. Calmon and
                  Himabindu Lakkaraju},
  editor       = {Amir Globersons and
                  Lester Mackey and
                  Danielle Belgrave and
                  Angela Fan and
                  Ulrich Paquet and
                  Jakub M. Tomczak and
                  Cheng Zhang},
  title        = {Interpreting {CLIP} with Sparse Linear Concept Embeddings ({SpLiCE})},
  booktitle    = {Advances in Neural Information Processing Systems 38: Annual Conference on Neural Information Processing Systems 2024, NeurIPS 2024},
  year         = {2024},
  url          = {http://papers.nips.cc/paper\_files/paper/2024/hash/996bef37d8a638f37bdfcac2789e835d-Abstract-Conference.html},
}

@inproceedings{ChoeOLCAS20,
  author       = {Junsuk Choe and
                  Seong Joon Oh and
                  Seungho Lee and
                  Sanghyuk Chun and
                  Zeynep Akata and
                  Hyunjung Shim},
  title        = {Evaluating Weakly Supervised Object Localization Methods Right},
  booktitle    = {2020 {IEEE/CVF} Conference on Computer Vision and Pattern Recognition, {CVPR} 2020},
  pages        = {3130--3139},
  publisher    = {Computer Vision Foundation / {IEEE}},
  year         = {2020},
  doi          = {10.1109/CVPR42600.2020.00320},
}

@inproceedings{RadfordKHRGASAM21,
  author       = {Alec Radford and
                  Jong Wook Kim and
                  Chris Hallacy and
                  Aditya Ramesh and
                  Gabriel Goh and
                  Sandhini Agarwal and
                  Girish Sastry and
                  Amanda Askell and
                  Pamela Mishkin and
                  Jack Clark and
                  Gretchen Krueger and
                  Ilya Sutskever},
  editor       = {Marina Meila and
                  Tong Zhang},
  title        = {Learning Transferable Visual Models From Natural Language Supervision},
  booktitle    = {Proceedings of the 38th International Conference on Machine Learning, {ICML} 2021},
  series       = {Proceedings of Machine Learning Research},
  volume       = {139},
  pages        = {8748--8763},
  publisher    = {{PMLR}},
  year         = {2021},
  url          = {http://proceedings.mlr.press/v139/radford21a.html},
}

@inproceedings{springsteinSRHK21,
  author       = {Matthias Springstein and
                  Stefanie Schneider and
                  Javad Rahnama and
                  Eyke H{\"{u}}llermeier and
                  Hubertus Kohle and
                  Ralph Ewerth},
  editor       = {Heng Tao Shen and
                  Yueting Zhuang and
                  John R. Smith and
                  Yang Yang and
                  Pablo C{\'{e}}sar and
                  Florian Metze and
                  Balakrishnan Prabhakaran},
  title        = {{iART}: {A} Search Engine for Art-Historical Images to Support Research in the Humanities},
  booktitle    = {{MM} '21: {ACM} Multimedia Conference},
  pages        = {2801--2803},
  publisher    = {{ACM}},
  year         = {2021},
  doi          = {10.1145/3474085.3478564},
}

@inproceedings{offertB2023,
  author      = {Fabian Offert and Peter Bell},
  editor      = {Anne Baillot and Toma Tasovac and Walter Scholger and Georg Vogeler},
  title       = {{imgs.ai}: {A} Deep Visual Search Engine for Digital Art History},
  booktitle   = {International Conference of the Alliance of Digital Humanities Organizations, {DH2022}},
  year        = {2023},
  doi         = {10.5281/zenodo.8107778},
}

@article{impettO2022,
  author       = {Leonardo Impett and 
                  Fabian Offert},
  title        = {There Is a Digital Art History},
  journal      = {Visual Resources},
  volume       = {38},
  number       = {2},
  pages        = {186--209},
  year         = {2022},
  doi          = {10.1080/01973762.2024.2362466}
}

@misc{birhanePK2021,
  author       = {Abeba Birhane and
                  Vinay Uday Prabhu and
                  Emmanuel Kahembwe},
  title        = {Multimodal Datasets: {M}isogyny, Pornography, and Malignant Stereotypes},
  year         = {2021},
  eprinttype   = {arXiv},
  eprint       = {2110.01963},
}

@inproceedings{benderGMS21,
  author       = {Emily M. Bender and
                  Timnit Gebru and
                  Angelina McMillan{-}Major and
                  Shmargaret Shmitchell},
  editor       = {Madeleine Clare Elish and
                  William Isaac and
                  Richard S. Zemel},
  title        = {On the Dangers of Stochastic Parrots: {C}an Language Models Be Too Big?},
  booktitle    = {FAccT '21: 2021 {ACM} Conference on Fairness, Accountability, and Transparency},
  pages        = {610--623},
  publisher    = {{ACM}},
  year         = {2021},
  doi          = {10.1145/3442188.3445922},
}

@inproceedings{liLSH23,
  author       = {Junnan Li and
                  Dongxu Li and
                  Silvio Savarese and
                  Steven C. H. Hoi},
  editor       = {Andreas Krause and
                  Emma Brunskill and
                  Kyunghyun Cho and
                  Barbara Engelhardt and
                  Sivan Sabato and
                  Jonathan Scarlett},
  title        = {{BLIP-2}: {B}ootstrapping Language-Image Pre-training with Frozen Image Encoders and Large Language Models},
  booktitle    = {International Conference on Machine Learning, {ICML} 2023},
  series       = {Proceedings of Machine Learning Research},
  volume       = {202},
  pages        = {19730--19742},
  publisher    = {{PMLR}},
  year         = {2023},
  url          = {https://proceedings.mlr.press/v202/li23q.html},
}

@inproceedings{zhaiWMSK0B22,
  author       = {Xiaohua Zhai and
                  Xiao Wang and
                  Basil Mustafa and
                  Andreas Steiner and
                  Daniel Keysers and
                  Alexander Kolesnikov and
                  Lucas Beyer},
  title        = {{LiT}: {Z}ero-Shot Transfer with Locked-image text Tuning},
  booktitle    = {{IEEE/CVF} Conference on Computer Vision and Pattern Recognition, {CVPR} 2022},
  pages        = {18102--18112},
  publisher    = {{IEEE}},
  year         = {2022},
  doi          = {10.1109/CVPR52688.2022.01759},
}

@misc{schuhmannVBKM2021,
  author       = {Christoph Schuhmann and
                  Richard Vencu and
                  Romain Beaumont and
                  Robert Kaczmarczyk and
                  Clayton Mullis and
                  Aarush Katta and
                  Theo Coombes and
                  Jenia Jitsev and
                  Aran Komatsuzaki},
  title        = {{LAION-400M}: {O}pen Dataset of {CLIP}-Filtered 400 Million Image-Text Pairs},
  year         = {2021},
  eprinttype   = {arXiv},
  eprint       = {2111.02114},
}

@article{hassijaCMSGHSSMH24,
  author       = {Vikas Hassija and
                  Vinay Chamola and
                  Atmesh Mahapatra and
                  Abhinandan Singal and
                  Divyansh Goel and
                  Kaizhu Huang and
                  Simone Scardapane and
                  Indro Spinelli and
                  Mufti Mahmud and
                  Amir Hussain},
  title        = {Interpreting Black-Box Models: {A} Review on Explainable Artificial Intelligence},
  journal      = {Cognitive Computation},
  volume       = {16},
  number       = {1},
  pages        = {45--74},
  year         = {2024},
  doi          = {10.1007/S12559-023-10179-8},
}

@inproceedings{speith22,
  author       = {Timo Speith},
  title        = {A Review of Taxonomies of Explainable Artificial Intelligence {(XAI)} Methods},
  booktitle    = {FAccT '22: 2022 {ACM} Conference on Fairness, Accountability, and Transparency},
  pages        = {2239--2250},
  publisher    = {{ACM}},
  year         = {2022},
  doi          = {10.1145/3531146.3534639},
}

@article{saeedO23,
  author       = {Waddah Saeed and
                  Christian W. Omlin},
  title        = {Explainable {AI} {(XAI):} {A} Systematic Meta-survey of Current Challenges and Future Opportunities},
  journal      = {Knowledge-Based Systems},
  volume       = {263},
  pages        = {110273},
  year         = {2023},
  doi          = {10.1016/J.KNOSYS.2023.110273},
}

@book{buchananS1984,
  author       = {Bruce G. Buchanan and Edward H. Shortliffe},
  title        = {Rule-based Expert Systems: {T}he {MYCIN} Experiments of the Stanford Heuristic Programming Project},
  publisher    = {Addison-Wesley},
  address      = {Reading, MA},
  year         = {1984},
}

@book{puppe1988,
  author       = {Frank Puppe},
  title        = {Einführung in Expertensysteme},
  publisher    = {Springer},
  address      = {Berlin, Heidelberg},
  year         = {1988},
}

@book{harmonMM1989,
  author       = {Paul Harmon and Rex Maus and William Morrissey},
  title        = {Expertensysteme: Werkzeuge und Anwendungen},
  publisher    = {Oldenbourg},
  address      = {München / Wien},
  year         = {1989},
}

@book{coyB1987,
  author       = {Wolfgang Coy and Lena Bonsiepen},
  title        = {Erfahrung und Berechnung: Kritik der Expertensystemtechnik},
  publisher    = {Springer},
  address      = {Berlin / Heidelberg},
  year         = {1987},
}

@article{kazmierczakBFF24,
  author       = {R{\'{e}}mi Kazmierczak and
                  Eloïse Berthier and
                  Goran Frehse and
                  Gianni Franchi},
  title        = {{CLIP-QDA}: {A}n Explainable Concept Bottleneck Model},
  journal      = {Transactions on Machine Learning Research},
  year         = {2024},
  url          = {https://openreview.net/pdf?id=jjmdiMiag7},
}

@inproceedings{abnarZ20,
  author       = {Samira Abnar and
                  Willem H. Zuidema},
  editor       = {Dan Jurafsky and
                  Joyce Chai and
                  Natalie Schluter and
                  Joel R. Tetreault},
  title        = {Quantifying Attention Flow in Transformers},
  booktitle    = {Proceedings of the 58th Annual Meeting of the Association for Computational Linguistics, {ACL} 2020},
  pages        = {4190--4197},
  publisher    = {Association for Computational Linguistics},
  year         = {2020},
  doi          = {10.18653/V1/2020.ACL-MAIN.385},
}

@inproceedings{jainW19,
  author       = {Sarthak Jain and
                  Byron C. Wallace},
  editor       = {Jill Burstein and
                  Christy Doran and
                  Thamar Solorio},
  title        = {Attention is not Explanation},
  booktitle    = {Proceedings of the 2019 Conference of the North American Chapter of the Association for Computational Linguistics: Human Language Technologies, {NAACL-HLT} 2019},
  pages        = {3543--3556},
  publisher    = {Association for Computational Linguistics},
  year         = {2019},
  doi          = {10.18653/V1/N19-1357},
}

@inproceedings{clip-dissect,
  author       = {Tuomas P. Oikarinen and
                  Tsui{-}Wei Weng},
  title        = {{CLIP-Dissect}: {A}utomatic Description of Neuron Representations in Deep
                  Vision Networks},
  booktitle    = {The 11th International Conference on Learning Representations, {ICLR} 2023},
  year         = {2023},
  url          = {https://openreview.net/pdf/a302e0072a6e15c8c0361c022bb9d3518f1a7127.pdf},
}


\onecolumn
\appendix
\renewcommand{\thefigure}{A.\arabic{figure}}
\setcounter{figure}{0}

\newcommand{\saliencymap}[2]{%
  \begin{subfigure}[t]{0.115\linewidth}
    \centering
    \includegraphics[width=\linewidth]{#1}
    \caption{#2}
  \end{subfigure}%
}

\begin{figure}[H]
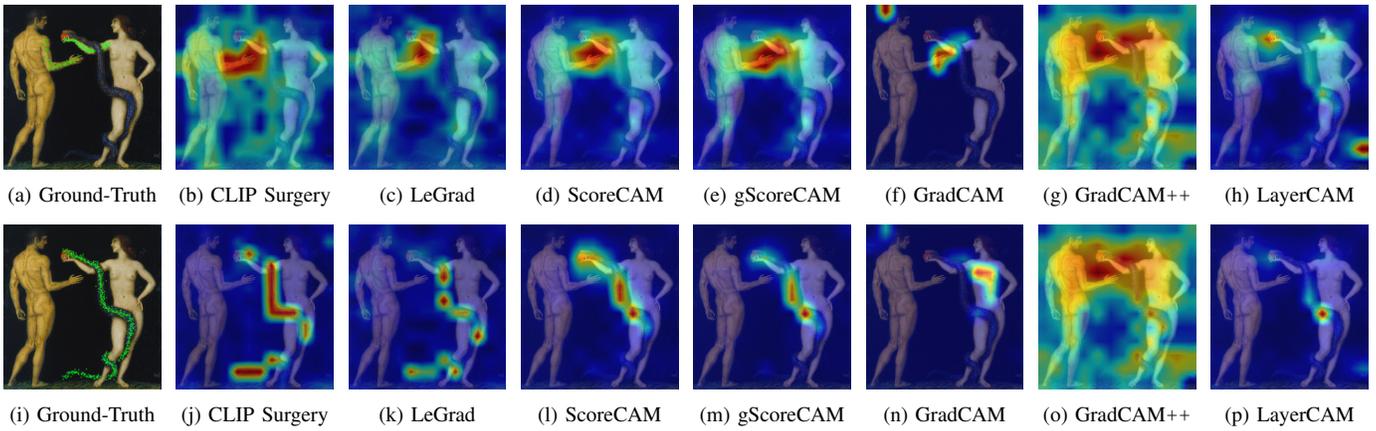

  \centering

  \foreach \img/\cap [count=\i] in {
    adam-and-eve.jpg_arm-outstretched.png/Ground-Truth,
    adam-and-eve.png_arm_outstretched_clip-surgery.jpg/CLIP Surgery,
    adam-and-eve.png_arm_outstretched_legrad.jpg/LeGrad,
    adam-and-eve.png_arm_outstretched_scorecam.jpg/ScoreCAM,
    adam-and-eve.png_arm_outstretched_gscorecam.jpg/gScoreCAM,
    adam-and-eve.png_arm_outstretched_gradcam.jpg/GradCAM,
    adam-and-eve.png_arm_outstretched_gradcam-plus.jpg/GradCAM\texttt{++},
    adam-and-eve.png_arm_outstretched_layercam.jpg/LayerCAM
  }{%
    \saliencymap{images/\img}{\cap}%
    \ifnum\i<8 \hfill\fi
  }
  \par\medskip
  \foreach \img/\cap [count=\j] in {
    adam-and-eve.jpg_snake.png/Ground-Truth,
    adam-and-eve.png_snake_clip-surgery.jpg/CLIP Surgery,
    adam-and-eve.png_snake_legrad.jpg/LeGrad,
    adam-and-eve.png_snake_scorecam.jpg/ScoreCAM,
    adam-and-eve.png_snake_gscorecam.jpg/gScoreCAM,
    adam-and-eve.png_snake_gradcam.jpg/GradCAM,
    adam-and-eve.png_snake_gradcam-plus.jpg/GradCAM\texttt{++},
    adam-and-eve.png_snake_layercam.jpg/LayerCAM
  }{%
    \saliencymap{images/\img}{\cap}%
    \ifnum\j<8 \hfill\fi
  }

  \caption{
    Ground-truth annotations and saliency maps for Franz von Stuck's \textit{Adam and Eve} (c.~1920), shown for the classes \enquote{arm outstretched} (\textit{top row}) and \enquote{snake} (\textit{bottom row}).
  }
  \label{fig:study-2-annotation-results-adam-and-eve}
\end{figure}

\begin{figure}[H]
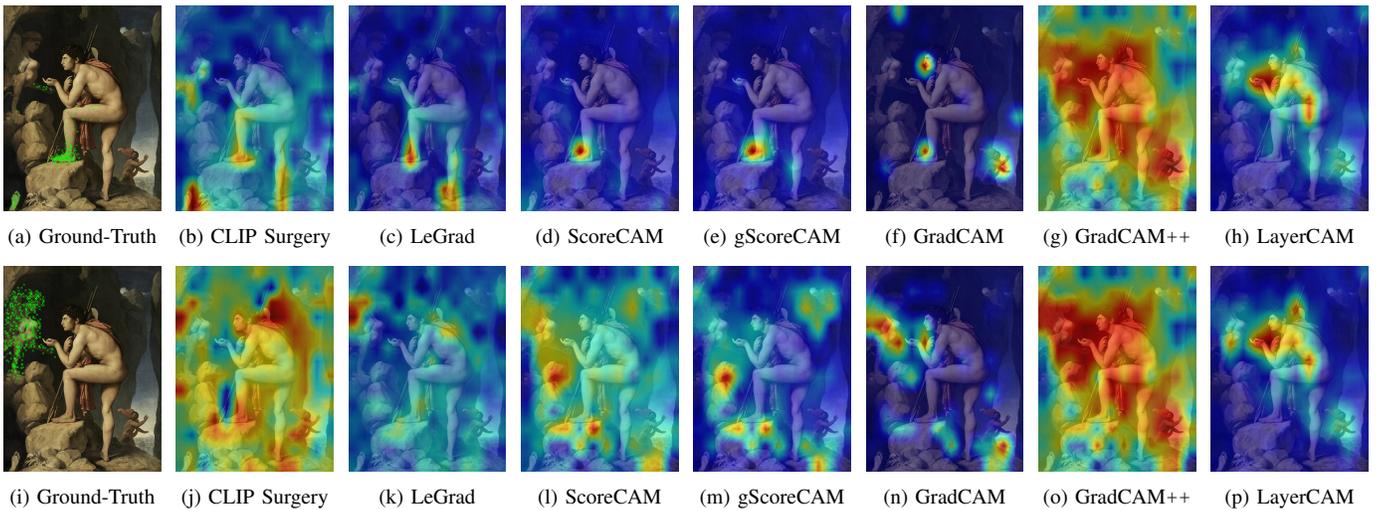

  \centering

  \foreach \img/\cap [count=\i] in {
    oedipus-and-the-sphinx.jpg_left-foot.png/Ground-Truth,
    oedipus-and-the-spinx.jpg_left_foot_clip-surgery.jpg/CLIP Surgery,
    oedipus-and-the-spinx.jpg_left_foot_legrad.jpg/LeGrad,
    oedipus-and-the-spinx.jpg_left_foot_scorecam.jpg/ScoreCAM,
    oedipus-and-the-spinx.jpg_left_foot_gscorecam.jpg/gScoreCAM,
    oedipus-and-the-spinx.jpg_left_foot_gradcam.jpg/GradCAM,
    oedipus-and-the-spinx.jpg_left_foot_gradcam-plus.jpg/GradCAM\texttt{++},
    oedipus-and-the-spinx.jpg_left_foot_layercam.jpg/LayerCAM
  }{%
    \saliencymap{images/\img}{\cap}%
    \ifnum\i<8 \hfill\fi
  }
  \par\medskip
  \foreach \img/\cap [count=\j] in {
    oedipus-and-the-sphinx.jpg_Sphinx.png/Ground-Truth,
    oedipus-and-the-spinx.jpg_Sphinx_clip-surgery.jpg/CLIP Surgery,
    oedipus-and-the-spinx.jpg_Sphinx_legrad.jpg/LeGrad,
    oedipus-and-the-spinx.jpg_Sphinx_scorecam.jpg/ScoreCAM,
    oedipus-and-the-spinx.jpg_Sphinx_gscorecam.jpg/gScoreCAM,
    oedipus-and-the-spinx.jpg_Sphinx_gradcam.jpg/GradCAM,
    oedipus-and-the-spinx.jpg_Sphinx_gradcam-plus.jpg/GradCAM\texttt{++},
    oedipus-and-the-spinx.jpg_Sphinx_layercam.jpg/LayerCAM
  }{%
    \saliencymap{images/\img}{\cap}%
    \ifnum\j<8 \hfill\fi
  }

  \caption{
    Ground-truth annotations and saliency maps for Jean-Auguste-Dominique Ingres's \textit{Oedipus and the Sphinx} (1808), shown for the classes \enquote{left foot} (\textit{top row}) and \enquote{Sphinx} (\textit{bottom row}).
  }
  \label{fig:study-2-annotation-results-oedipus-and-the-sphinx}
\end{figure}

\begin{figure}[H]
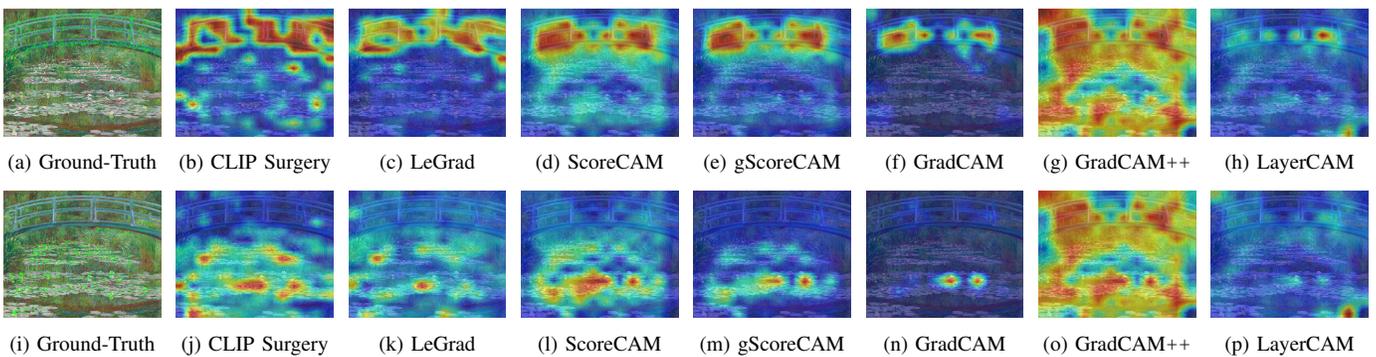

  \centering

  \foreach \img/\cap [count=\i] in {
    japanese-footbridge.jpg_bridge.png/Ground-Truth,
    japanese-footbridge.jpg_bridge_clip-surgery.jpg/CLIP Surgery,
    japanese-footbridge.jpg_bridge_legrad.jpg/LeGrad,
    japanese-footbridge.jpg_bridge_scorecam.jpg/ScoreCAM,
    japanese-footbridge.jpg_bridge_gscorecam.jpg/gScoreCAM,
    japanese-footbridge.jpg_bridge_gradcam.jpg/GradCAM,
    japanese-footbridge.jpg_bridge_gradcam-plus.jpg/GradCAM\texttt{++},
    japanese-footbridge.jpg_bridge_layercam.jpg/LayerCAM
  }{%
    \saliencymap{images/\img}{\cap}%
    \ifnum\i<8 \hfill\fi
  }
  \par\medskip
  \foreach \img/\cap [count=\j] in {
    japanese-footbridge.jpg_flower.png/Ground-Truth,
    japanese-footbridge.jpg_flower_clip-surgery.jpg/CLIP Surgery,
    japanese-footbridge.jpg_flower_legrad.jpg/LeGrad,
    japanese-footbridge.jpg_flower_scorecam.jpg/ScoreCAM,
    japanese-footbridge.jpg_flower_gscorecam.jpg/gScoreCAM,
    japanese-footbridge.jpg_flower_gradcam.jpg/GradCAM,
    japanese-footbridge.jpg_flower_gradcam-plus.jpg/GradCAM\texttt{++},
    japanese-footbridge.jpg_flower_layercam.jpg/LayerCAM
  }{%
    \saliencymap{images/\img}{\cap}%
    \ifnum\j<8 \hfill\fi
  }

  \caption{
    Ground-truth annotations and saliency maps for Claude Monet's \textit{Japanese Footbridge} (1899), shown for the classes \enquote{bridge} (\textit{top row}) and \enquote{flower} (\textit{bottom row}).
  }
  \label{fig:study-2-annotation-results-japanese-footbridge}
\end{figure}

\begin{figure}[H]
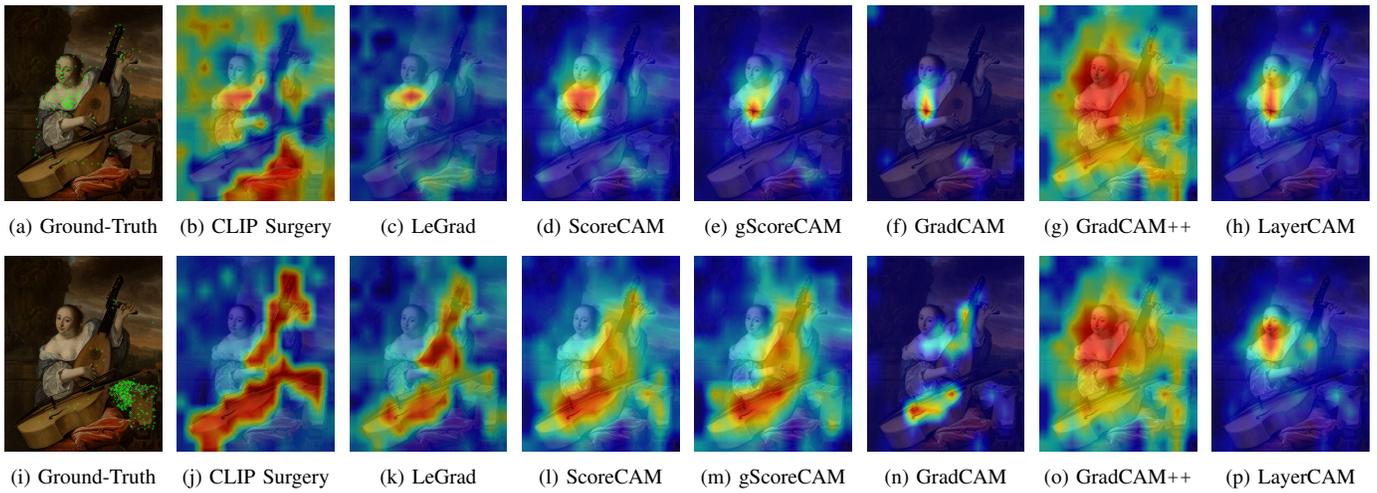

  \centering

  \foreach \img/\cap [count=\i] in {
    the-musician.jpg_lustful.png/Ground-Truth,
    the-musician.jpg_lustful_clip-surgery.jpg/CLIP Surgery,
    the-musician.jpg_lustful_legrad.jpg/LeGrad,
    the-musician.jpg_lustful_scorecam.jpg/ScoreCAM,
    the-musician.jpg_lustful_gscorecam.jpg/gScoreCAM,
    the-musician.jpg_lustful_gradcam.jpg/GradCAM,
    the-musician.jpg_lustful_gradcam-plus.jpg/GradCAM\texttt{++},
    the-musician.jpg_lustful_layercam.jpg/LayerCAM
  }{%
    \saliencymap{images/\img}{\cap}%
    \ifnum\i<8 \hfill\fi
  }
  \par\medskip
  \foreach \img/\cap [count=\j] in {
    the-musician.jpg_sheet-music.png/Ground-Truth,
    the-musician.jpg_sheet_music_clip-surgery.jpg/CLIP Surgery,
    the-musician.jpg_sheet_music_legrad.jpg/LeGrad,
    the-musician.jpg_sheet_music_scorecam.jpg/ScoreCAM,
    the-musician.jpg_sheet_music_gscorecam.jpg/gScoreCAM,
    the-musician.jpg_sheet_music_gradcam.jpg/GradCAM,
    the-musician.jpg_sheet_music_gradcam-plus.jpg/GradCAM\texttt{++},
    the-musician.jpg_sheet_music_layercam.jpg/LayerCAM
  }{%
    \saliencymap{images/\img}{\cap}%
    \ifnum\j<8 \hfill\fi
  }

  \caption{
    Ground-truth annotations and saliency maps for Bartholomeus van der Helst's \textit{The Musician} (1662), shown for the classes \enquote{lustful} (\textit{top row}) and \enquote{sheet music} (\textit{bottom row}).
  }
  \label{fig:study-2-annotation-results-the-musician}
\end{figure}

\begin{figure}[H]
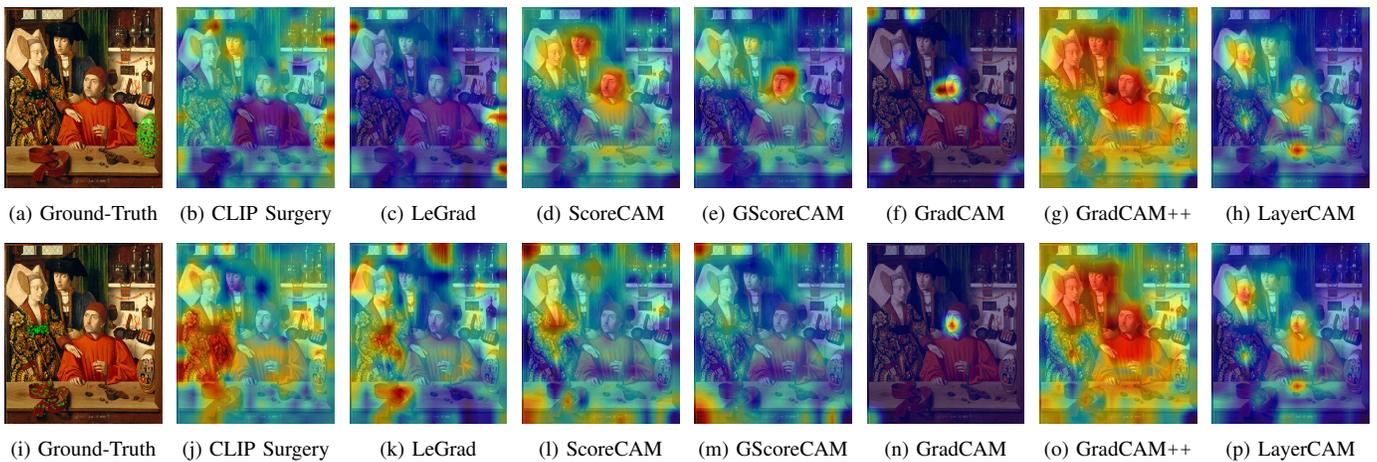

  \centering

  \foreach \img/\cap [count=\i] in {
    a-goldsmith-in-his-shop.jpg_convex-mirror.png/Ground-Truth,
    a-goldsmith-in-his-shop.jpg_convex_mirror_clip-surgery.jpg/CLIP Surgery,
    a-goldsmith-in-his-shop.jpg_convex_mirror_legrad.jpg/LeGrad,
    a-goldsmith-in-his-shop.jpg_convex_mirror_scorecam.jpg/ScoreCAM,
    a-goldsmith-in-his-shop.jpg_convex_mirror_gscorecam.jpg/GScoreCAM,
    a-goldsmith-in-his-shop.jpg_convex_mirror_gradcam.jpg/GradCAM,
    a-goldsmith-in-his-shop.jpg_convex_mirror_gradcam-plus.jpg/GradCAM\texttt{++},
    a-goldsmith-in-his-shop.jpg_convex_mirror_layercam.jpg/LayerCAM
  }{%
    \saliencymap{images/\img}{\cap}%
    \ifnum\i<8 \hfill\fi
  }
  \par\medskip

  \foreach \img/\cap [count=\j] in {
    a-goldsmith-in-his-shop.jpg_girdle.png/Ground-Truth,
    a-goldsmith-in-his-shop.jpg_girdle_clip-surgery.jpg/CLIP Surgery,
    a-goldsmith-in-his-shop.jpg_girdle_legrad.jpg/LeGrad,
    a-goldsmith-in-his-shop.jpg_girdle_scorecam.jpg/ScoreCAM,
    a-goldsmith-in-his-shop.jpg_girdle_gscorecam.jpg/GScoreCAM,
    a-goldsmith-in-his-shop.jpg_girdle_gradcam.jpg/GradCAM,
    a-goldsmith-in-his-shop.jpg_girdle_gradcam-plus.jpg/GradCAM\texttt{++},
    a-goldsmith-in-his-shop.jpg_girdle_layercam.jpg/LayerCAM
  }{%
    \saliencymap{images/\img}{\cap}%
    \ifnum\j<8 \hfill\fi
  }

  \caption{
    Ground-truth annotations and saliency maps for Petrus Christus's \textit{A Goldsmith in his Shop} (1449), shown for the classes \enquote{convex mirror} (\textit{top row}) and \enquote{girdle} (\textit{bottom row}).
  }
  \label{fig:study-2-annotation-results-a-goldsmith-in-his-shop}
\end{figure}

\begin{figure}[H]
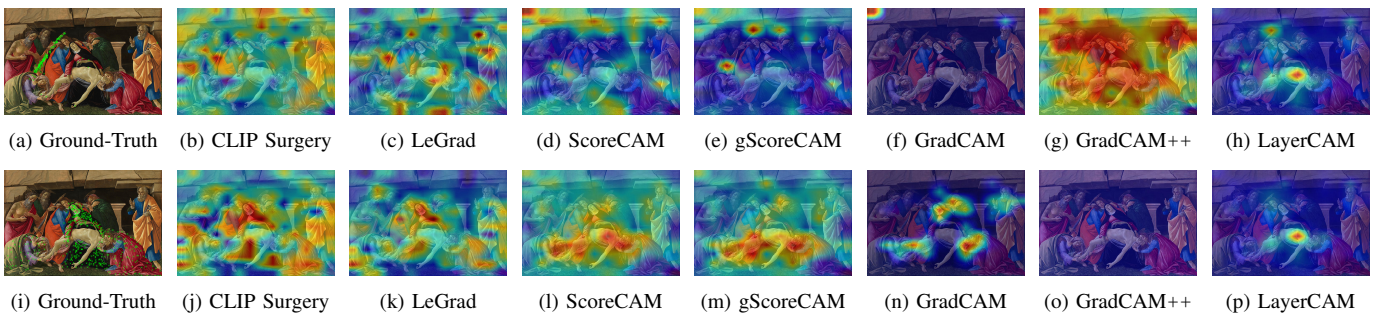

  \centering

  \foreach \img/\cap [count=\i] in {
    the-lamentation.jpg_sword.png/Ground-Truth,
    the-lamentation.jpg_sword_clip-surgery.jpg/CLIP Surgery,
    the-lamentation.jpg_sword_legrad.jpg/LeGrad,
    the-lamentation.jpg_sword_scorecam.jpg/ScoreCAM,
    the-lamentation.jpg_sword_gscorecam.jpg/gScoreCAM,
    the-lamentation.jpg_sword_gradcam.jpg/GradCAM,
    the-lamentation.jpg_sword_gradcam-plus.jpg/GradCAM\texttt{++},
    the-lamentation.jpg_sword_layercam.jpg/LayerCAM
  }{%
    \saliencymap{images/\img}{\cap}%
    \ifnum\i<8 \hfill\fi
  }
  \par\medskip
  \foreach \img/\cap [count=\j] in {
    the-lamentation.jpg_Virgin-Mary.png/Ground-Truth,
    the-lamentation.jpg_Virgin_Mary_clip-surgery.jpg/CLIP Surgery,
    the-lamentation.jpg_Virgin_Mary_legrad.jpg/LeGrad,
    the-lamentation.jpg_Virgin_Mary_scorecam.jpg/ScoreCAM,
    the-lamentation.jpg_Virgin_Mary_gscorecam.jpg/gScoreCAM,
    the-lamentation.jpg_Virgin_Mary_gradcam.jpg/GradCAM,
    the-lamentation.jpg_Virgin_Mary_gradcam-plus.jpg/GradCAM\texttt{++},
    the-lamentation.jpg_Virgin_Mary_layercam.jpg/LayerCAM
  }{%
    \saliencymap{images/\img}{\cap}%
    \ifnum\j<8 \hfill\fi
  }

  \caption{
    Ground-truth annotations and saliency maps for Sandro Botticelli's \textit{The Lamentation} (c.~1490), shown for the classes \enquote{sword} (\textit{top row}) and \enquote{Virgin Mary} (\textit{bottom row}).
  }
  \label{fig:study-2-annotation-results-the-lamentation}
\end{figure}

\begin{figure}[H]
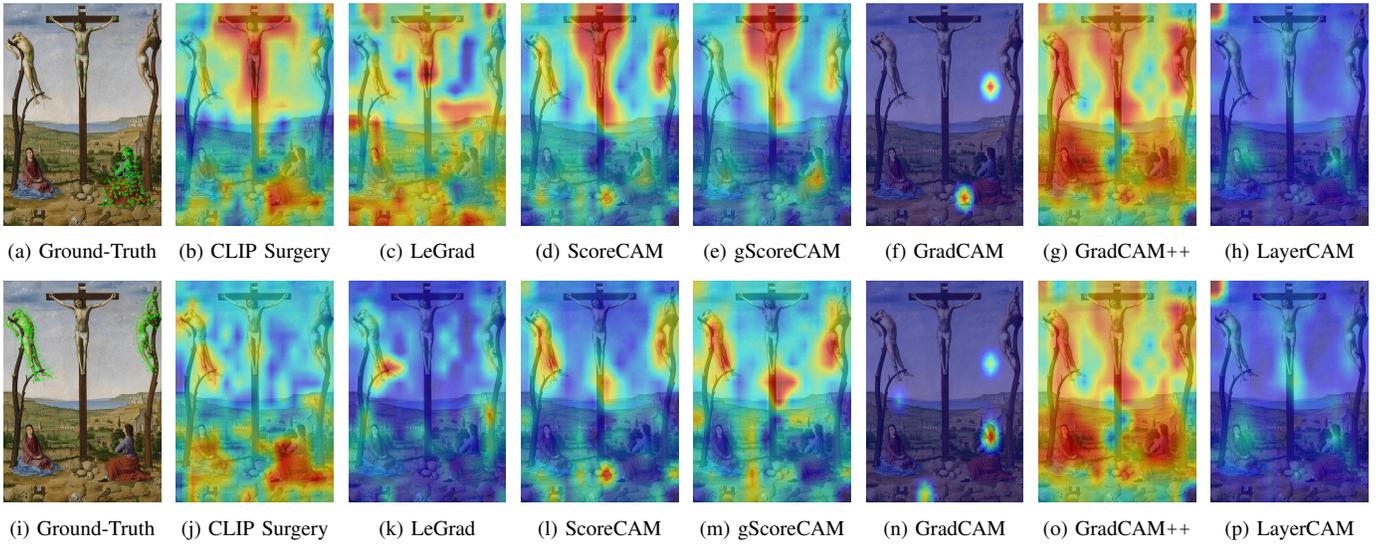

  \centering

  \foreach \img/\cap [count=\i] in {
    calvary.jpg_John.png/Ground-Truth,
    calvary.jpg_John_clip-surgery.jpg/CLIP Surgery,
    calvary.jpg_John_legrad.jpg/LeGrad,
    calvary.jpg_John_scorecam.jpg/ScoreCAM,
    calvary.jpg_John_gscorecam.jpg/gScoreCAM,
    calvary.jpg_John_gradcam.jpg/GradCAM,
    calvary.jpg_John_gradcam-plus.jpg/GradCAM\texttt{++},
    calvary.jpg_John_layercam.jpg/LayerCAM
  }{%
    \saliencymap{images/\img}{\cap}%
    \ifnum\i<8 \hfill\fi
  }
  \par\medskip
  \foreach \img/\cap [count=\j] in {
    calvary.jpg_thief.png/Ground-Truth,
    calvary.jpg_thief_clip-surgery.jpg/CLIP Surgery,
    calvary.jpg_thief_legrad.jpg/LeGrad,
    calvary.jpg_thief_scorecam.jpg/ScoreCAM,
    calvary.jpg_thief_gscorecam.jpg/gScoreCAM,
    calvary.jpg_thief_gradcam.jpg/GradCAM,
    calvary.jpg_thief_gradcam-plus.jpg/GradCAM\texttt{++},
    calvary.jpg_thief_layercam.jpg/LayerCAM
  }{%
    \saliencymap{images/\img}{\cap}%
    \ifnum\j<8 \hfill\fi
  }

  \caption{
    Ground-truth annotations and saliency maps for Antonello da Messina's \textit{Calvery} (1475), shown for the classes \enquote{John} (\textit{top row}) and \enquote{thief} (\textit{bottom row}).
  }
  \label{fig:study-2-annotation-results-calvary}
\end{figure}

\end{document}